\pdfoutput=1

\documentclass[11pt]{article}

\usepackage[]{acl}

\usepackage{microtype}
\usepackage[utf8]{inputenc}
\usepackage[T1]{fontenc}
\usepackage{times}
\usepackage{latexsym}
\usepackage{amsfonts}
\usepackage{amsmath}
\usepackage{amssymb}
\usepackage{amsthm}
\usepackage{relsize}
\usepackage{xspace}
\usepackage{placeins} 

\usepackage[group-separator={,}]{siunitx}

\usepackage{tikz}
\usetikzlibrary{bayesnet}
\usetikzlibrary{arrows}
\usepackage{color}
\usepackage{graphicx}
\usepackage{caption}
\usepackage{subcaption}
\usetikzlibrary{backgrounds}

\usepackage{hhline}
\usepackage{xcolor}
\usepackage{colortbl}
\definecolor{Grey}{RGB}{210,210,210}

\def\calX{{\mathcal{X}}}

\usepackage{enumitem}

\usepackage{algorithm}
\usepackage{algpseudocode}

\usepackage{multirow}

\usepackage{amsmath,amsfonts,bm}
\usepackage{dsfont}

\def\eqref#1{equation~\ref{#1}}

\def\1{\bm{1}}

\DeclareMathAlphabet{\mathsfit}{\encodingdefault}{\sfdefault}{m}{sl}
\SetMathAlphabet{\mathsfit}{bold}{\encodingdefault}{\sfdefault}{bx}{n}

\newcommand{\R}{\mathbb{R}}

\newcommand{\softmax}{\mathrm{softmax}}

\usepackage{bbm}

\usepackage{todonotes}

\makeatletter
\newcommand*\iftodonotes{\if@todonotes@disabled\expandafter\@secondoftwo\else\expandafter\@firstoftwo\fi}  
\makeatother

\usepackage{tipa}
\newcommand{\ethz}{\boldsymbol{\texttt{E}}}
\newcommand{\google}{\boldsymbol{\texttt{G}}}

\newcommand*{\escape}[1]{\texttt{\textbackslash#1}}

\newcommand{\setQuote}[1]{``{#1}''}

\newcommand{\co}{{contrastive objective}\xspace}
\newcommand{\CO}{\textsc{CO}\xspace}

\newcommand{\gptNeoSmall}{\textsc{GPT-Neo 125M}\xspace}
\newcommand{\gptNeo}{\textsc{GPT-Neo}\xspace}
\newcommand{\gptThree}{\textsc{GPT-3}\xspace}

\newcommand{\pile}{\textsc{Pile}\xspace}

\definecolor{memRGB}{RGB}{245, 15, 15}
\definecolor{nonmemRGB}{RGB}{2, 26, 255}
\definecolor{perturbmemRGB}{RGB}{165, 15, 245}

\newcommand{\memColor}[1]{{\color{memRGB} #1}}
\newcommand{\nonmemColor}[1]{{\color{nonmemRGB} #1}}
\newcommand{\perturbmemColor}[1]{{\color{perturbmemRGB} #1}}

\newcommand{\memPara}{\memColor{memorized paragraph}\xspace}

\newcommand{\perturbmemPara}{\perturbmemColor{perturbed memorized paragraph}\xspace}

\newcommand{\memParas}{\memColor{memorized paragraphs}\xspace}
\newcommand{\nonmemParas}{\nonmemColor{non-memorized paragraphs}\xspace}
\newcommand{\perturbmemParas}{\perturbmemColor{perturbed memorized paragraphs}\xspace}

\newcommand{\MP}{\memColor{\textsc{MP}}\xspace}
\newcommand{\NMP}{\nonmemColor{\textsc{NMP}}\xspace}

\newcommand{\MPs}{\memColor{\textsc{MPs}}\xspace}
\newcommand{\NMPs}{\nonmemColor{\textsc{NMPs}}\xspace}
\newcommand{\PMPs}{\perturbmemColor{\textsc{PMPs}}\xspace}

\newcommand{\Xmem}{\memColor{\mathcal{X}^{\textsc{M}}}}
\newcommand{\xmem}{\memColor{\boldsymbol{x}_{n}^{\textsc{M}}}}

\newcommand{\Xnonmem}{\nonmemColor{\mathcal{X}^{\textsc{NM}}}}
\newcommand{\xnonmem}{\nonmemColor{\boldsymbol{x}_{N}^{\textsc{NM}}}}

\newcommand{\xperturbmem}{\perturbmemColor{\boldsymbol{x}_{n}^{\textsc{M}}}}

\newcommand{\KLDiv}{\mathcal{D}_\mathrm{KL}}
\newcommand{\NLLLoss}{\mathcal{L}_\mathrm{NLL}}

\newcommand{\params}{\boldsymbol{\theta}} 
\newcommand{\paramsZero}{\boldsymbol{\theta_{0}}}

\newcommand{\paramGradAttr}{\Delta \theta_{l,c}}

\newcommand{\NLL}{\textrm{NLL}\xspace}
\newcommand{\EM}{\textrm{EM}\xspace}

\usepackage{cleveref}
\crefname{section}{\S}{\S\S}
\Crefname{section}{\S}{\S\S}
\crefname{table}{Tab.}{}
\crefname{figure}{Fig.}{}
\crefname{algorithm}{Algorithm}{}
\crefname{equation}{Eq.}{}
\crefname{appendix}{App.}{}
\crefname{thm}{Theorem}{}
\crefname{prop}{Proposition}{}
\crefname{cor}{Corollary}{}
\crefname{observation}{Observation}{}
\crefname{assumption}{Assumption}{}
\crefformat{section}{\S#2#1#3}

\makeatletter
\newcommand\footnoteref[1]{\protected@xdef\@thefnmark{\ref{#1}}\@footnotemark}
\makeatother

\usepackage{times}
\usepackage{latexsym}

\usepackage[T1]{fontenc}

\usepackage[utf8]{inputenc}

\usepackage{microtype}

\usepackage{inconsolata}

\title{Localizing Paragraph Memorization in Language Models}

\author{
Niklas Stoehr$^{\ethz}$\thanks{~ Work done while at Google} \qquad Mitchell Gordon$^{\google}$ \qquad Chiyuan Zhang$^{\google}$ \qquad Owen Lewis$^{\google}$\\
\\
$^{\ethz}$ETH Z{\"u}rich \qquad $^{\google}$Google\\
\footnotesize \href{mailto:niklas.stoehr@inf.ethz.ch}{\texttt{niklas.stoehr@inf.ethz.ch}} \qquad \{\href{mailto:mitchellgordon@google.com}{\texttt{mitchellgordon}}, \href{mailto:chiyuan@google.com}{\texttt{chiyuan}}, \href{mailto:lewiso@google.com}{\texttt{lewiso}}\}\texttt{@google.com}
}

\begin{document}
\maketitle

\begin{abstract}
Can we localize the weights and mechanisms used by a language model to memorize and recite entire paragraphs of its training data? In this paper, we show that while memorization is spread across multiple layers and model components, gradients of memorized paragraphs have a distinguishable spatial pattern, being larger in lower model layers than gradients of non-memorized examples. Moreover, the memorized examples can be unlearned by fine-tuning only the high-gradient weights. We localize a low-layer attention head that appears to be especially involved in paragraph memorization. This head is predominantly focusing its attention on distinctive, rare tokens that are least frequent in a corpus-level unigram distribution. Next, we study how localized memorization is across the tokens in the prefix by perturbing tokens and measuring the caused change in the decoding. A few distinctive tokens early in a prefix can often corrupt the entire continuation. Overall, memorized continuations are not only harder to unlearn, but also to corrupt than non-memorized ones. 
\end{abstract}

\let\thefootnote\relax\footnotetext{Code and data: \href{https://github.com/googleinterns/localizing-paragraph-memorization}{github.com/googleinterns/localizing-paragraph-memorization}}

\section{Introduction}

\begin{figure}[t!]
 \centering
 \includegraphics[width=1.0\linewidth]{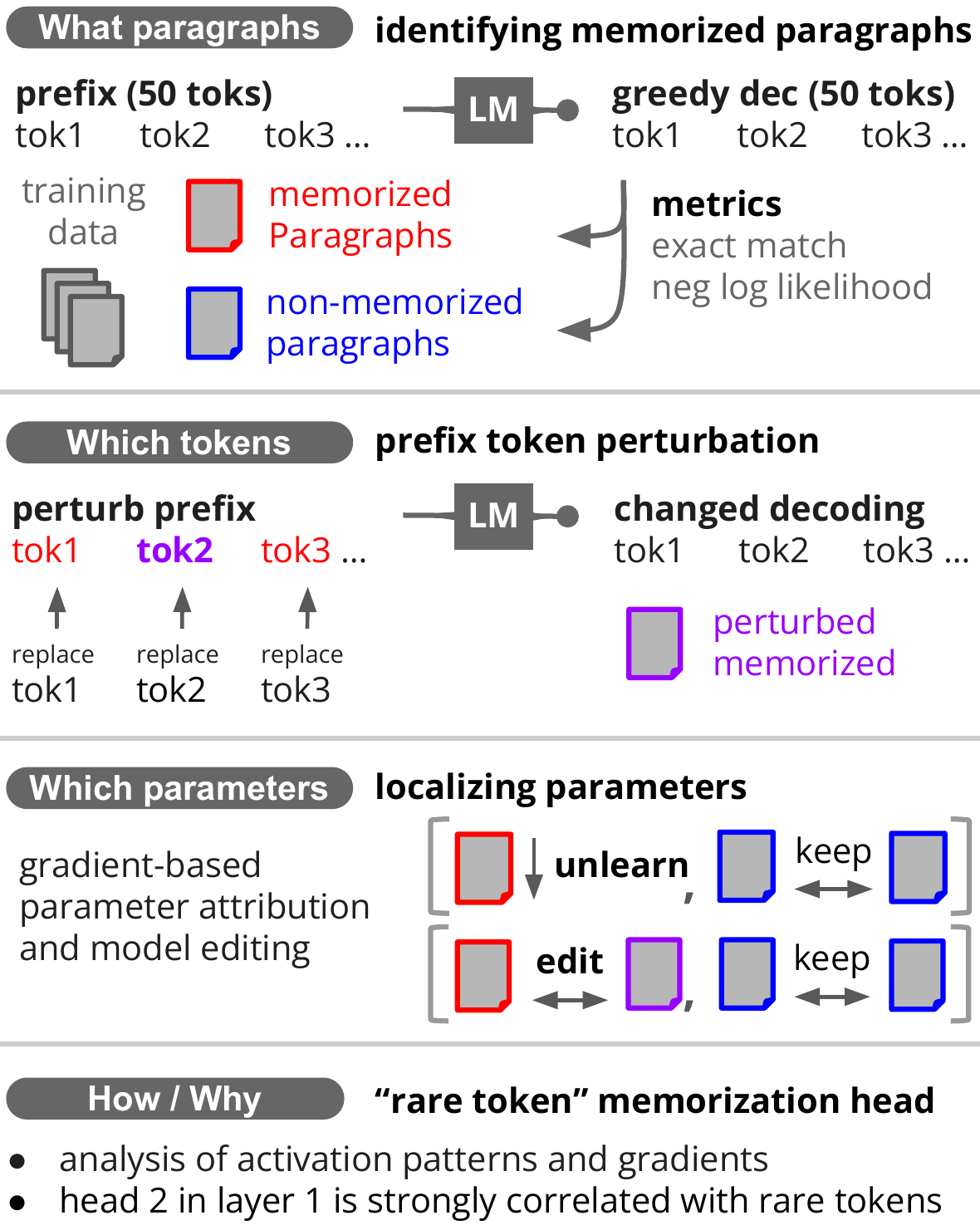} 
\caption{We interpret language models with respect to their capability to memorize \num{100}-token paragraphs from the training data. Using sets of \memColor{memorized}, \nonmemColor{non-memorized} as well as \perturbmemColor{perturbed memorized} paragraphs, we study parameter and activation gradients, activation patterns as well as unlearning and editing objectives to identify an influential \setQuote{memorization head}.
}
\label{fig:overview}
\end{figure}

Some language models are able to emit gigabytes of full-length paragraphs from their training data \citep{carlini_extracting_2020, carlini_quantifying_2022, mccoy_how_2023, haviv_understanding_2023, nasr_scalable_2023, new_york_times_one_2023}. These memorized paragraphs must thus be represented somewhere in the model weights \citep{nasr_scalable_2023}. We take steps towards localizing these weights and internal mechanisms that are involved in the memorization of paragraphs. Specifically, we study in detail the open-weight model \gptNeoSmall \citep{gao_pile_2021} which has been trained on the publicly available dataset the \pile.

As a first step, we identify paragraphs that are memorized by a language model. We use the term \setQuote{paragraph} for any sequence of \num{100} tokens. A paragraph is regarded as memorized if, given a prefix of \num{50} tokens, the model's greedy decoding of the next \num{50} tokens exactly matches the true paragraph continuation. We publish the memorized paragraphs alongside our code.

We use our dataset of memorized and non-memorized paragraphs to identify differences in how they are processed by the model. To this end, we measure the effect that perturbing individual tokens in a paragraph's prefix has on the model's memorization. We find that \setQuote{memorization triggers} can sometimes be localized to few, distinctive tokens very early in the prefix. Moreover, corrupting memorized paragraphs is, on average, more difficult than non-memorized paragraphs. The perturbed prefix continuations of previously memorized paragraphs are mostly still semantically and syntactically valid and can be regarded as alternative paraphrases.

These experiments localize \setQuote{when} memorized information is accessed throughout the paragraph. To understand \setQuote{where} this information may be stored, we turn to the model's parameters which are shared across all token positions. We find that parameter gradients flow indeed differently for memorized and non-memorized paragraphs. To better isolate these gradient differences, we adapt a contrastive objective from prior work \citep{maini_can_2023} that seeks to reduce the likelihood of memorized paragraphs while leaving non-memorized paragraphs unchanged. This objective has the additional advantage that it can be used to (sparsely) fine-tune the model: we upgrade only those parameters that we have previously localized and validate that our localization does in fact inform editing \citep{hase_does_2023}. In particular, we experiment with two fine-tuning objectives, one that \setQuote{unlearns} and one that \setQuote{edits} memorized paragraphs into their perturbed alternatives. We find that unlearning is easier than editing, and it is often difficult to leave non-memorized paragraphs unchanged. 

While memorization is spread across multiple layers and components of the model, there is one model component that is standing out: attention head 2 in layer 1. Analyzing activation gradients and attention patterns, we qualitatively and quantitatively show that this head attends predominantly to distinctive, or rare tokens in the long tail of the unigram token distribution. We include additional experiments with activation patching and activation gradients in the appendix.

\section{Related Work}
\label{sec:related_work}

This paper connects three lines of work on language models: memorization, interpretability and editing. 

\paragraph{Memorization in Language Models.}

Our work builds upon \citet{carlini_quantifying_2022}, who quantify which and how many paragraphs from the training data are memorized by open-source language models such as \gptNeo \citep{gao_pile_2021}. This setup, where an adversary attempts to efficiently recover memorized training data, has been extensively studied on language models~\citep{carlini_extracting_2020, zhang_counterfactual_2021, nasr_scalable_2023}. Other related work focuses on n-gram novelty versus copying from the training data \citep{mccoy_how_2023}. \citet{hartmann_sok_2023} and \citet{zheng_empirical_2022} provide surveys on types of memorization and their risks with respect to alignment, privacy and copyright. Importantly, we do not study any differences in model behavior on paragraphs within vs outside of the training data. This is another important privacy-related aspect known as Membership Inference Attack \citep{hu_membership_2021, mattern_membership_2023, shi_detecting_2023}.

\paragraph{Language Model Interpretability.}

Beyond identifying \setQuote{what} training set paragraphs are memorized, we are interested in interpreting \setQuote{how} a model does so. \citet{chang_localization_2023} test whether different localization methods agree when localizing memorization in language models. The studied methods include brute-force zeroing out of model weights, learning a mask to prune weights and removing weights based on gradient attribution. In this work, we predominantly focus on gradient-based attribution \citep{sundararajan_axiomatic_2017, du_generalizing_2023}, but also draw inspirations from activation patching~\citep{meng_locating_2022, geva_dissecting_2023} which aims at localizing the memorization of few-token facts instead of paragraphs. Existing interpretability work \citep{chang_localization_2023, haviv_understanding_2023} studies shorter memorized text spans such as idioms, URLs or quotes, for which memorization may have a different definition than for \num{100}-token paragraphs. In \cref{sec:localizing_parameters}, we borrow methods for gradient-based attribution using a contrastive objective from \citet{maini_can_2023}. While their work focuses on memorizing atypical training set examples in image classification, we adapt their methods to memorization of paragraphs in language models. Related to our \setQuote{memorization head} in \cref{sec:memorization_head}, \citet{yu_characterizing_2023} identify a \setQuote{memory head} which however plays a widely different role. It down-weights geographic knowledge in in-context QA tasks. 

\begin{figure}[t]
 \centering
 \includegraphics[width=1.0\linewidth]{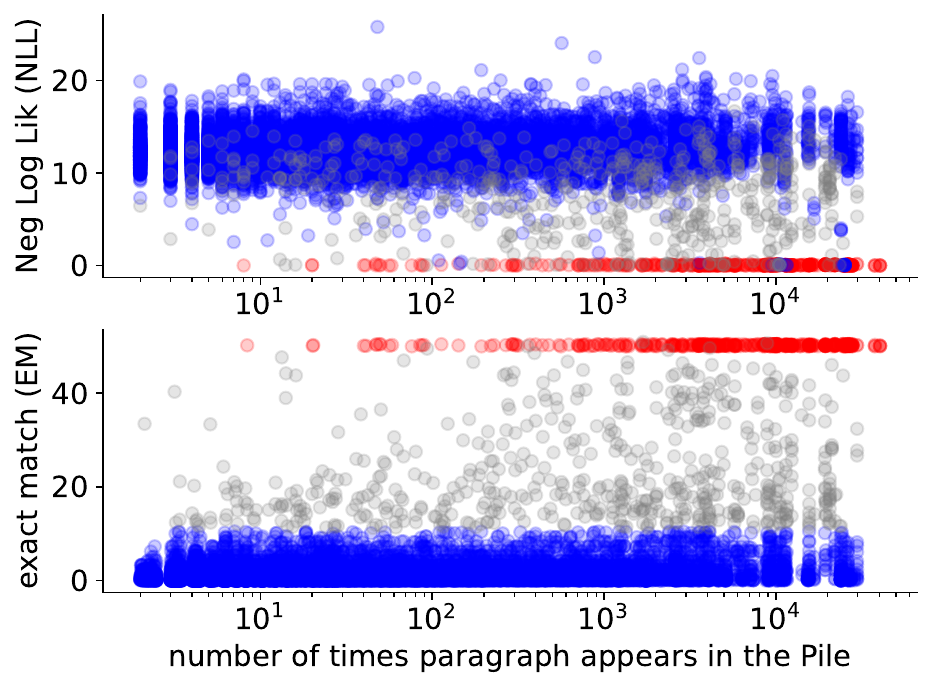} 
\caption{Splitting paragraphs of the \pile into \memParas and \nonmemParas based on \gptNeoSmall. We present the model with paragraph prefixes of length \num{50} tokens, greedy decode the next \num{50} tokens and evaluate the generation in terms of negative log-likelihood (\NLL) and exact match (\EM).}
\label{fig:nll_em}
\end{figure}

\paragraph{Model Editing and Unlearning.}

\citet{hase_does_2023} ask whether \setQuote{localization inform[s] editing} and led us to confirm our localization of relevant model parameters by fine-tuning only those parameters in an unlearning and model editing setting. Similar to their findings, we observe that memorization components are spread out across layers while patching-based methods in \cref{sec:patching} point to other components. Our model editing setup in \cref{sec:editing} is similar to \citet{eldan_whos_2023}, who find alternative paraphrases of facts that they use to fine-tune a model. Related areas of study are language model watermarking \citep{kirchenbauer_watermark_2023} and grokking \citep{power_grokking_2022}.

\begin{figure*}[t]
 \centering
 \includegraphics[width=1.0\linewidth]{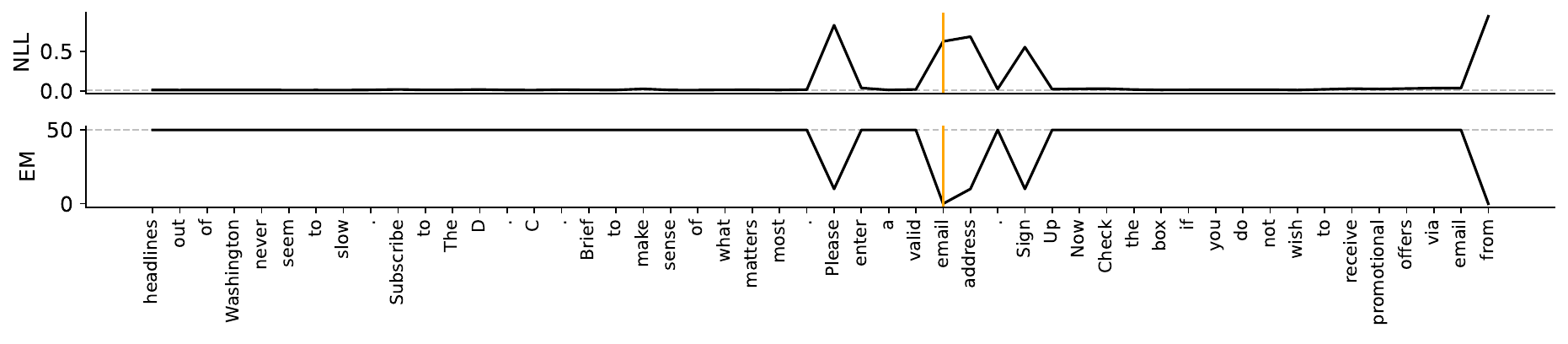} 
 \includegraphics[width=1.0\linewidth]{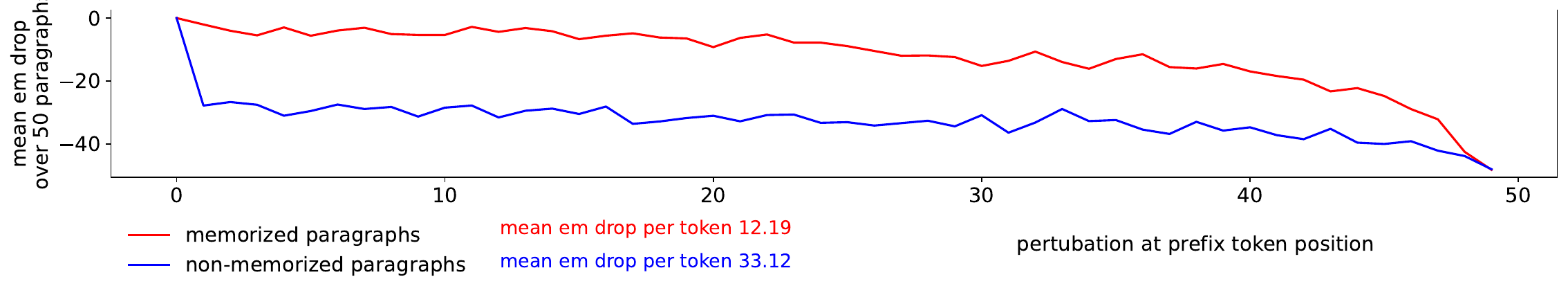} 
\caption{\textbf{[top]} The plot shows the effect of perturbing tokens in the prefix (shown) on the model's generation (not shown) in terms of the negative log-likelihood (\NLL) and exact match (\EM). Changing the single token \setQuote{email} into a random other token causes the \EM to drop by \num{45}, even though \setQuote{email} is about \num{20} tokens before the generated part.
\textbf{[bottom]} Perturbing tokens in the \memParas has, on average, less impact in exact match drop (\EM) in the model's generation, than perturbing tokens in the \nonmemParas.}
\label{fig:prefix_perturb}
\end{figure*}

\section{Identifying Memorized Paragraphs}
\label{sec:identifying}

\subsection{Open-Source Model and Training Set}
\label{sec:model_data}

\paragraph{\gptNeoSmall.} We seek to zoom in on a selected model to study its specific memorization behavior in detail. All presented methodology can however be transferred to any open-weight model. The \gptNeo family of models \citep{gao_pile_2021} is intended to be the open-weight counterpart to the \gptThree model \citep{brown_language_2020} in terms of model architecture. \gptNeo models are trained on a publicly available dataset, the \pile \citep{gao_pile_2021}, which allows checking model generations against its training data. As such, they have been studied extensively with respect to how much they memorize \citep{carlini_quantifying_2022, nasr_scalable_2023}. While these studies found that bigger model variants tend to memorize more, the smallest variant, \gptNeoSmall, still exhibits extensive memorization behavior with an easier-to-study computational footprint. After all, when interpreting models at the level of individual weights, smaller models are easier to visualize and analyze.

\begin{figure*}[t]
 \centering
 \includegraphics[width=1.0\linewidth]{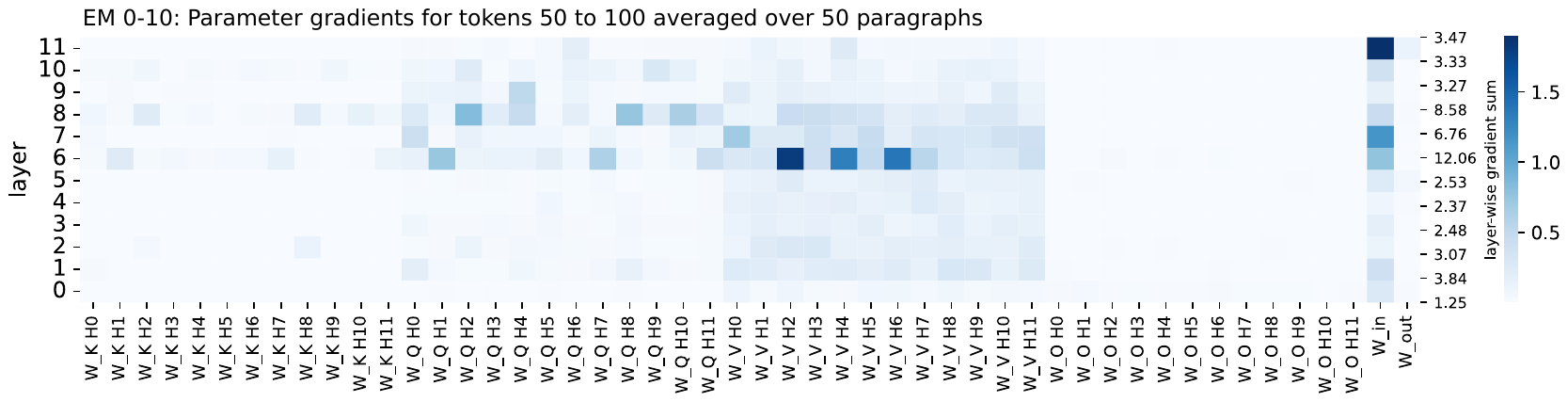} 
  \includegraphics[width=1.0\linewidth]{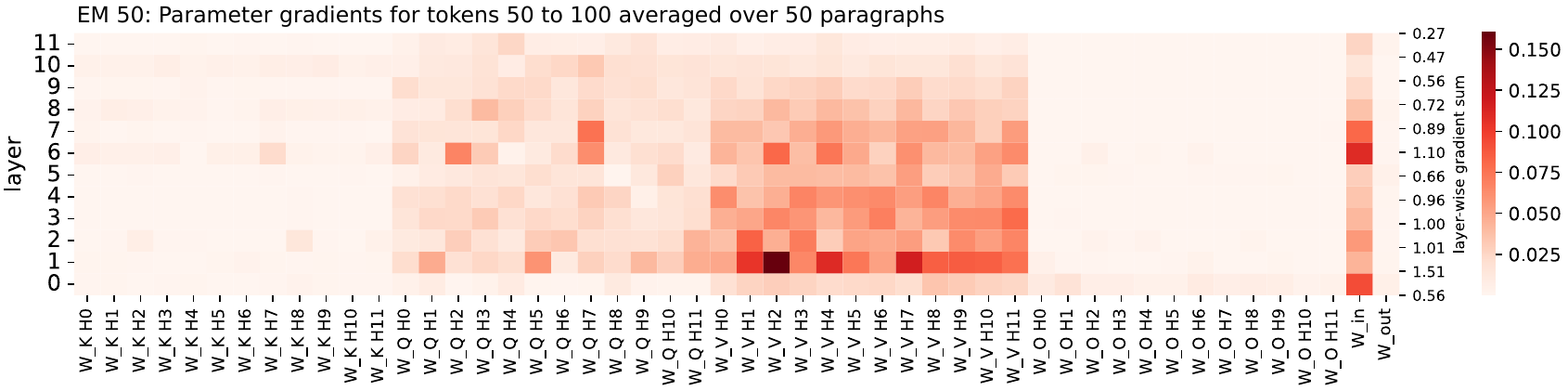} 
  \includegraphics[width=1.0\linewidth]{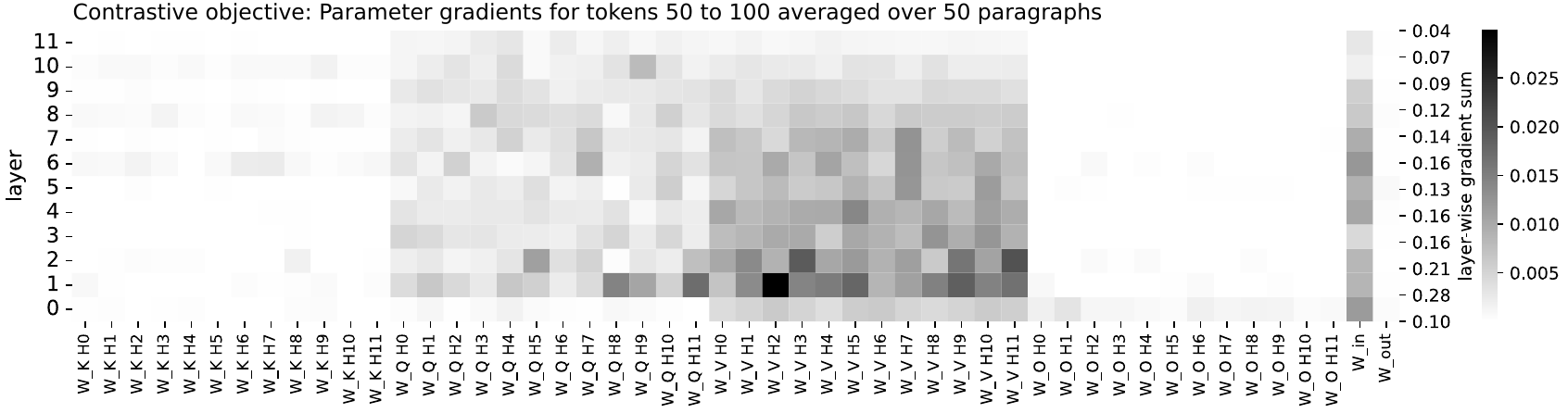}
\includegraphics[width=1.0\linewidth]{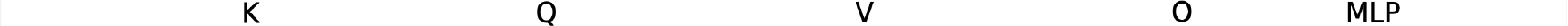}
\caption{\textbf{[top and center]} While memorization appears to be spread across multiple layers, we observe systemically different parameter gradients for \memColor{memorized} and \nonmemColor{non-memorized paragraphs}. The former is associated with lower absolute gradients in lower layers of the model. \textbf{[bottom]} Parameter gradient attribution scores for the \co (\cref{eq:contrast_objective}).The value matrix ($\texttt{W\_V}$) of attention head 2 in layer 1 appears to be strongly involved.}
\label{fig:params_gradients}
\end{figure*}

\paragraph{The \pile.} \gptNeoSmall was trained on the \pile \citep{gao_pile_2021}, an aggregation of \num{22} different datasets. It comprises \num{825}GB of English text and code. For this study, we consider a post-processed \href{https://github.com/ethz-spylab/lm_memorization_data/tree/main/data}{\num{570}GB subset} of the \pile provided by \citet{carlini_quantifying_2022}. This subset contains \num{10000} randomly sampled, unique \num{100}-token paragraphs and the count how frequently they occur in the training set. We perform pre-processing steps to find a diverse set of paragraphs as detailed in \cref{sec:preprocessing}. This leaves us with \num{13450} paragraphs of which the most frequent one occurs \num{40382} times in the \pile.

\subsection{Memorization Metrics and Data Split}
\label{sec:metrics}

We split the \num{13450} \pile paragraphs $\calX$ into a set of \memParas (\MP) and \nonmemParas (\NMP) which are disjoint subsets $\calX = \Xmem \cup \Xnonmem$. To this end, we consider the exact match (\EM) of the model's greedy decoding in an \setQuote{extractable memorization} setting \citep{nasr_scalable_2023}. We also take the negative log-likelihood (\NLL) into consideration. 

\paragraph{Exact Match (\EM).}

Exact match (\EM) is the number of greedily decoded tokens that exactly match the tokens in the ground truth training set paragraph until the first mismatch. Since the continuations are \num{50} tokens long, $\EM = 50$ is the maximum value. 

\paragraph{Negative Log-Likelihood (\NLL).}

Under a model with parameters $\params$, the negative log-likelihood for a batch of $N$ paragraphs $x_{N,I}$ that are each $I$ tokens long is given by $\NLLLoss(\boldsymbol{x}_{N,I}; \params) = \frac{1}{N} \sum_{n}^{N} \Big( - \frac{1}{I} \sum_{i}^{I} \log p (x_{n,i} \mid \boldsymbol{x}_{n,0:i-1}; \params) \Big)$. All paragraphs studied in this work are $I=100$ tokens long of which the first \num{50} tokens are the prefix. We compute the \NLL only on the last \num{50} (generated) tokens and omit the token position index $i$ for simplicity in the following.

\paragraph{Memorized Paragraphs.} \cref{fig:nll_em} shows the \NLL and \EM results for all paragraphs. We select the \num{442} paragraphs with $\EM = 50$ as the memorized set which is clearly distinct, both in terms of \NLL and \EM, from the other paragraphs. We provide an overview of some exemplary \MPs in App. \cref{tab:memorized_paragraphs}. Setting boundaries for a non-memorized set is less clear, but we choose the \num{12422} paragraphs with $0 \leq \EM \leq 10$. Similar to the $\EM = 50$ paragraphs, those paragraphs form a distinctive cluster in \cref{fig:nll_em}. While there is high overlap when splitting based on \NLL and \EM, we observe that splitting based on \NLL yields less diverse, even more code-based examples since those generally have lower \NLL. We hypothesize this is because there are less \setQuote{second-best} paraphrases / alternatives for code.

\section{Prefix Token Perturbation}
\label{sec:token_perturbation}

Where in the paragraph do interventions disrupt memorization the most? We study this question by perturbing every token in the prefix, one token at a time, by replacing it with a random token from the vocabulary. For every \num{50}-token prefix with a single perturbed token, we then use the language model to obtain a greedy decoding of the next \num{50} tokens. We measure the change in the decoding caused by the perturbation in terms of \NLL and \EM as shown at the top of \cref{fig:prefix_perturb}. For different \MPs, we often see that a few, distinctive tokens, even at early positions in the prefix, lead to a drop in \EM of up to \num{45}. 

In \cref{fig:prefix_perturb} at the bottom, we zoom in on this finding by computing the mean \EM drop per prefix token position over \num{50} \MPs and \NMPs. As expected, the closer the token to the decoded tokens (later in the prefix), the more impact the token has on the decoding. Interestingly, \NMPs are, on average, easier perturbed than \MPs. This may be hint at one property of memorization---\MPs seem more \setQuote{baked} into the model while \NMPs with generally lower likelihood can easily \setQuote{slip off} into equally likely paraphrases.   

If a single token is able to divert the model's continuation of an \MP, what does this continuation look like? The examples in \cref{tab:perturbed_continuations} in the appendix demonstrate that the model's generations are syntactically and semantically mostly valid. In the following, we refer to those continuations based off a perturbed prefix as \perturbmemParas (\PMPs). \PMPs can be seen as admissible paraphrases of \MPs.

\section{Localizing Parameters}
\label{sec:localizing_parameters}

We investigate if there are any systematic differences in how the model internally processes our sets of \MPs and \NMPs. While we previously looked at token positions, we now turn to an analysis of model parameters which are shared across all token positions. Taking a simplified view, the model parameters are of shape $\params \in \R^{L \times C \times D^{*}}$, where $\{l\}_{0}^{L}$ indexes into the model's \num{12} layers, also known as Transformer blocks \citep{vaswani_attention_2017}. For \gptNeoSmall, each layer $l$ consists of $C = 50$ model component types, $c \in \{ \texttt{W\_K H0, W\_K H1,} \ldots \}$. The attention mechanism is comprised of \num{12} attention heads, each consisting of a key $\texttt{W\_K}$, query $\texttt{W\_Q}$, value $\texttt{W\_V}$, and output $\texttt{W\_O}$ matrix. The multi-layer perceptron (MLP) block per layer consists of the input $\texttt{W\_in}$ and output matrix $\texttt{W\_out}$. The layers and model components are shown on the Y and X axis in \cref{fig:params_gradients} respectively. $D^{*}$ refers to the vector dimension, i.e., the number of weights which varies for each model component, thus, \setQuote{D star} for simplicity. 

\begin{figure*}[t]
 \centering
 \includegraphics[width=1.0\linewidth]{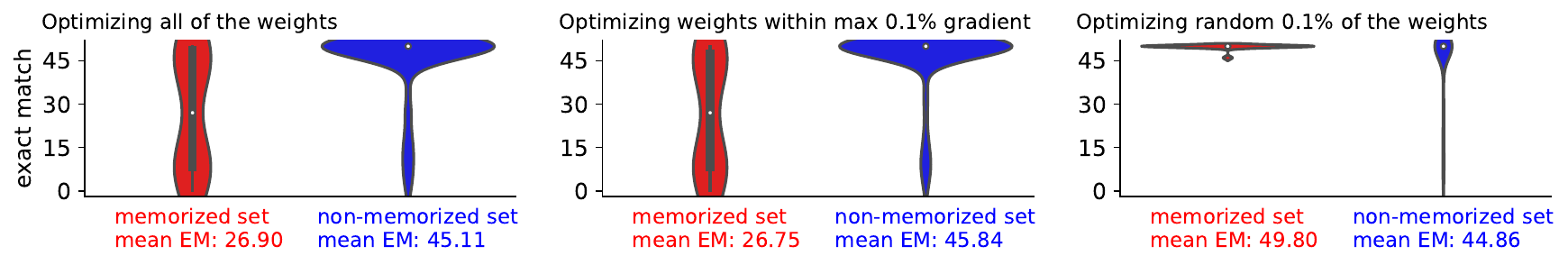} 
 \includegraphics[width=1.0\linewidth]{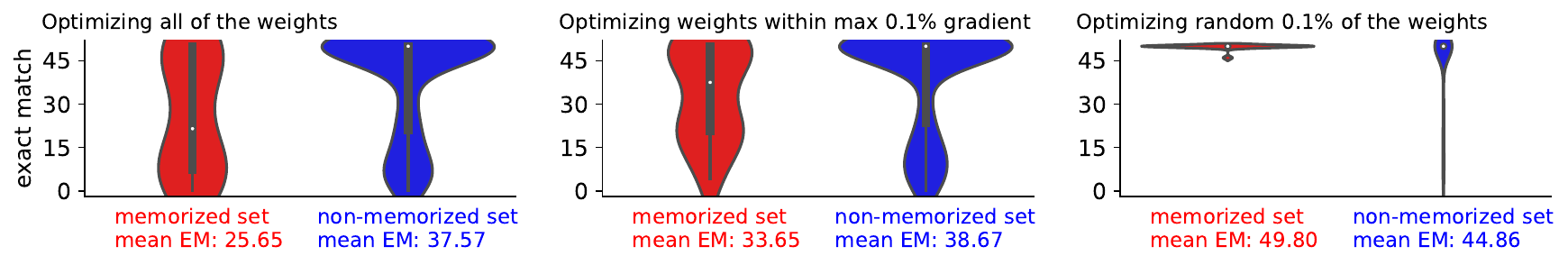} 
\caption{\textbf{[top]} To test whether our localization also informs editing, we optimize all model parameters based on the \co (\cref{eq:contrast_objective}), only the \num{0.1}\% weights with the maximum gradient and a random sample of weights. Result shows that sparsely fine-tuning only the max gradient weights causes the most unlearning in \MPs and the least in \NMPs. \textbf{[bottom]} Instead of unlearning \MPs, we consider an editing objective (\cref{eq:model_editing}) to overwrite \MPs using \PMPs. While sparse optimization of only the max gradient weights appears to be similarly effective as training all weights, editing is overall more difficult than unlearning.
}
\label{fig:editing}
\end{figure*}

\subsection{Gradient-based Parameter Attribution}
\label{sec:parameter_gradients}

We feed a batch of paragraphs to the language model and compute the \NLL loss $\NLLLoss$ for tokens \num{50} to \num{100}, i.e., the generation of the model given the prefix. We then compute the parameter gradients $\Delta \params \in \R^{L \times C \times D^{*}}$ with respect to the loss:
\begin{align}
    \Delta \params = \frac{\partial \NLLLoss(\boldsymbol{x}_{N}; \params)}{\partial  \params}
    \label{eq:parameter_gradients}
\end{align}
To obtain a \emph{parameter gradient attribution score} $\paramGradAttr$, we consider the absolute gradient value for all individual weights and choose the maximum value per layer $l$ and component $c$:
\begin{align}
    \paramGradAttr = \underset{d}{\mathrm{max}} \big( \lvert \{ \Delta \theta_{l,c,d} \}_{d}^{D^{*}} \rvert \big)
    \label{eq:gradient_attribution}
\end{align}
In \cref{fig:params_gradients}, we present the mean parameter gradient attribution scores for a batch of \num{50} \MPs and, separately, a batch of \num{50} \NMPs. We observe clear differences between the attribution scores: first of all, but less surprisingly, the gradients for the \NMPs are larger since those are less likely under the model \citep{shi_detecting_2023}. More surprising are the clear differences with respect to layers: there is more gradient flow for \MPs in lower layers, for both attention and MLP components, which is in line with \citet{haviv_understanding_2023}. In fact, we observe a smooth shift in gradient patterns when evaluating \setQuote{partly memorized} paragraphs with $10 \leq \EM \leq 50$ as displayed in App. \cref{fig:params_gradient_4}.

\subsection{Contrastive Objective}
\label{sec:contrastive_objective}

Inspired by \citet{chang_localization_2023}'s localization method, we combine \MPs and \NMPs in a contrastive objective. The objective is to change memorized continuations of \MPs while preserving the model's continuations of \NMPs, which translates into the following \co (\CO):
\begin{align}
    \label{eq:contrast_objective}
    \CO_{\downarrow}(\xmem, \xnonmem; \params) &=  -\NLLLoss(\xmem; \params) \\ \nonumber
    &+ \KLDiv \big( (\xnonmem; \params), (\xnonmem; \paramsZero) \big)  
\end{align}
The \CO increases the \NLL of an individual \MP $\xmem$ and decreases the KL divergence $\KLDiv$ from the model's original continuations of a batch of $N$ \NMPs $\xnonmem$. This set of \NMPs can be seen as a \setQuote{control} set that ensures the model remains as much as possible unaltered. We denote $\paramsZero$ as the model's original parameters which are excluded (detached) from the gradient computation. To study the removal of multiple \MPs, we recompute the \CO over \num{50} different \MPs and randomly sampled batches of \NMPs and aggregate all gradient computations. We rely on \href{https://neelnanda-io.github.io/TransformerLens/}{\texttt{TransformerLens}}
\citep{nanda_transformerlenslibrary_2023} for the implementation of this and the following experiments. We disable gradient computation on the model components \setQuote{\texttt{embed}}, \setQuote{\texttt{pos\_embed}}, \setQuote{\texttt{unembed}} and all bias terms. As shown in \cref{fig:params_gradients}, the parameter gradient attribution scores yield by the \co reveal similar patterns to those observed in \cref{fig:params_gradients}. Most importantly, in both settings, the value matrix ($\texttt{W\_V}$) of attention head 2 in layer 1 is most salient. 

\subsection{Sparse Unlearning and Editing}
\label{sec:editing}
 
Instead of computing gradients to only obtain attribution scores, we may also update the model parameters based on the gradients in an optimization setting to satisfy the \co (\CO) in \cref{eq:contrast_objective}. This can help us find further evidence that the localized parameters are meaningful for memorization \citep{hase_does_2023}. 

\paragraph{Unlearning \MPs.}

We compute the gradients of all parameters with respect to the \CO and mask out all parameters that are not within the maximum \num{0.1} \% of all absolute gradient values. We keep this mask while taking \num{10} gradient steps using the Adam optimizer \citep{kingma_adam_2015} which can be seen as a form of sparse fine-tuning. We compare this setting against optimizing all of the weights and masking a random \num{0.1} \% of the weights as shown in \cref{fig:editing}. While the goal is to bring down the \EM of \MPs from formerly \num{50} to \num{0}, the \EM of the model's original continuation of the \NMPs should remain unchanged ($\EM = 50$). We find that the result between optimizing all weights versus only the \num{0.1}\% max gradient weights does not worsen. To the contrary, there is even more drop in \EM on the \MPs and less drop on the \NMPs. Moreover, optimizing a randomly selected \num{0.1}\% of weights does not achieve the desired result at all.

\begin{figure*}[t]
 \centering
\includegraphics[width=1.0\linewidth]{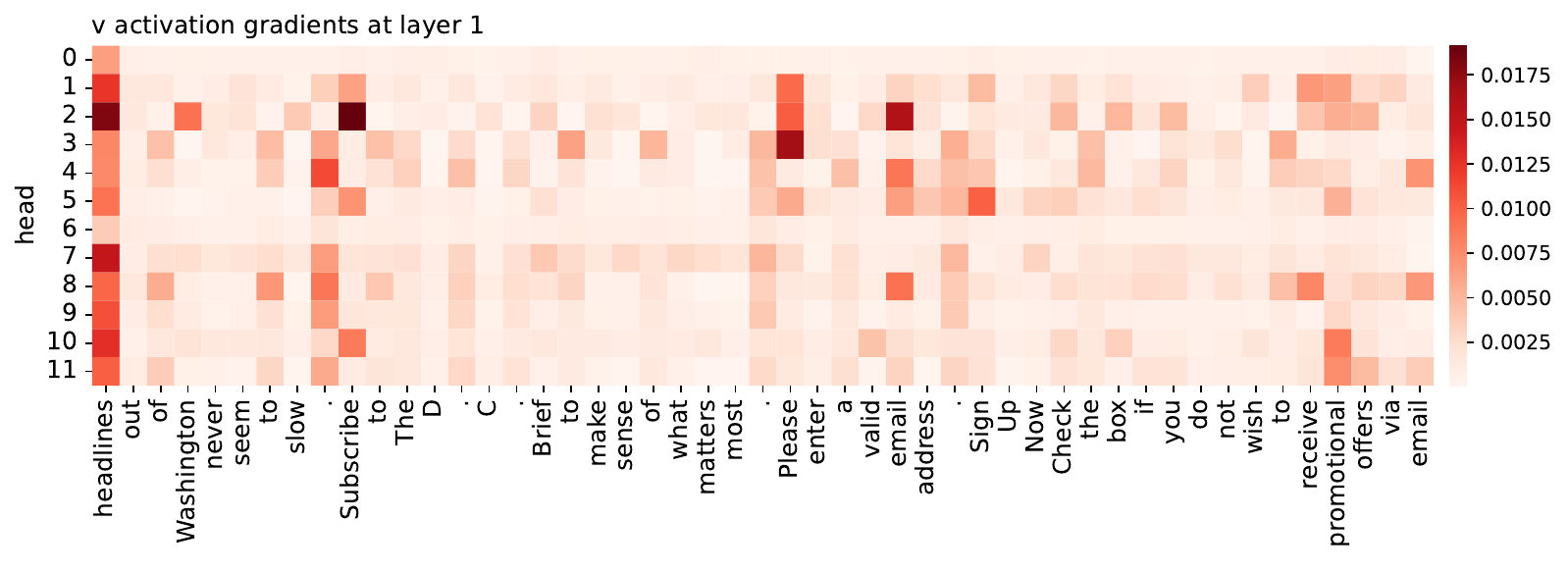} 
 \includegraphics[width=1.0\linewidth]{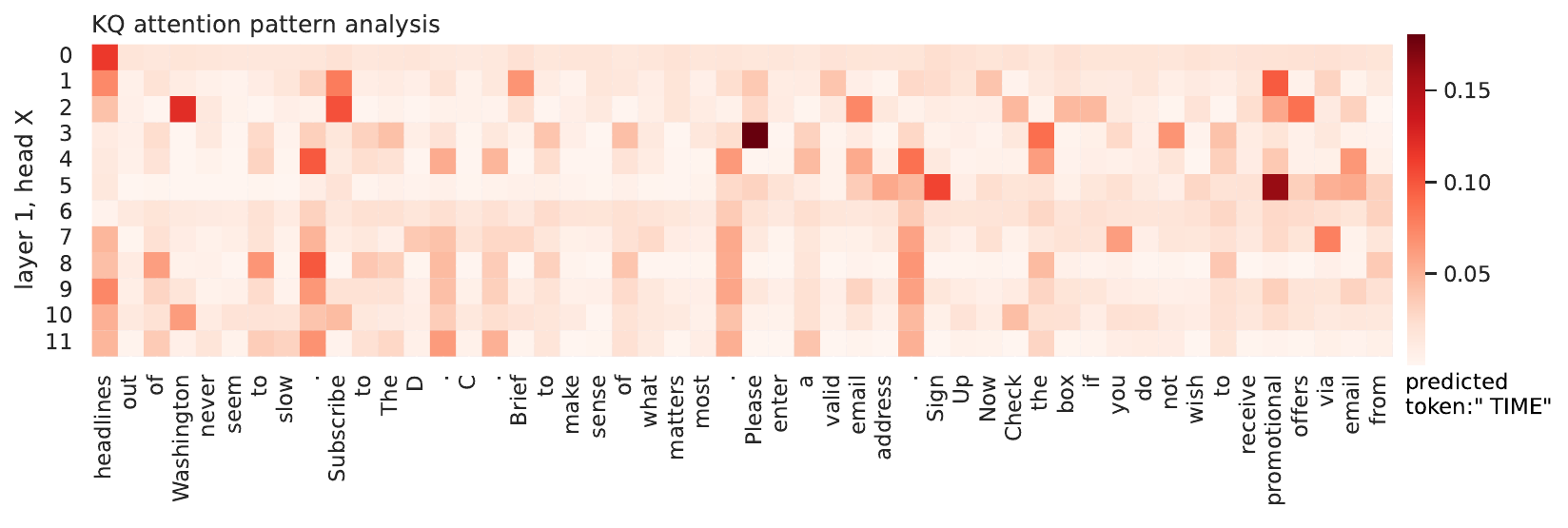} 
\caption{\textbf{[Top]} Value activation gradients on layer 1. \textbf{[Bottom]} KQ attention on layer 1. We find that head 2 shows similar attention patterns in both, \textbf{[Top]} and \textbf{[Bottom]}: more distinctive tokens such as \setQuote{Washington}, \setQuote{Subscribe} or \setQuote{email} are more influential and are often the ones causing most perturbation to memorized paragraphs (\cref{sec:token_perturbation}).}
\label{fig:attention_pattern_l1}
\end{figure*}

\paragraph{Editing \MPs into \PMPs.}

Instead of \setQuote{unlearning} \MPs, we make an effort to edit them into \PMPs with a modified \co:
\begin{align}
    \label{eq:model_editing}
    \CO_{\leftrightarrow}(\xperturbmem, \xnonmem; \params) &=  +\NLLLoss(\xperturbmem; \params) \\ \nonumber
    &+ \KLDiv \big( (\xnonmem; \params), (\xnonmem; \paramsZero) \big)  
\end{align}
Instead of increasing the \NLL on \MPs $\xmem$, we are now decreasing the \NLL on \PMPs $\xperturbmem$ to make their alternative continuations more likely. The editing results for \num{10} optimization steps is presented in \cref{fig:editing}. Again, optimizing only a masked \num{0.1}\% of high gradient weights performs equally well to optimizing all weights. Comparing results however suggests that unlearning is easier than editing. A common finding from perturbing the prefix (\cref{fig:prefix_perturb}), unlearning and editing \MPs (\cref{fig:editing}) is that it is indeed difficult to remove \MPs while leaving \NMPs unchanged.   

\section{Memorization Head L1H2}
\label{sec:memorization_head}

In \cref{sec:localizing_parameters}, different analysis methods point to the same model component, the value matrix of attention head \num{2} in layer \num{1}. This is in line with \citet{haviv_understanding_2023} who find that memorized tokens are promoted in lower layers and it motivates us to study the role of this specific head in more detail.

\subsection{Activation Gradients}
\label{sec:activation_gradients}

Instead of computing gradients with respect to parameters as in \cref{eq:parameter_gradients}, we now compute gradients with respect to activations $\boldsymbol{h} \in \R^{L \times C \times I \times D^{*}}$: 
\begin{align}
    \Delta \boldsymbol{h} = \frac{\partial \NLLLoss(\boldsymbol{x}_{N}; \boldsymbol{h})}{\partial  \boldsymbol{h}}
    \label{eq:activation_gradients}
\end{align}
As before, we consider absolute gradients and max-pool over the (hidden) dimension $D^{*}$ to obtain attribution scores $\Delta h_{l,c,i}$ per layer $l$, model component $c$ and token position $i$. \cref{fig:attention_pattern_l1} [top] shows the value activation attribution scores for layer 1 for an exemplary \MP. Again, head 2 appears to be particularly active and somewhat anti-correlated with the other heads. For instance, head's 2 gradient attribution is large for the tokens \setQuote{Subscribe} or \setQuote{Washington}, and not for their neighboring tokens \setQuote{.} or \setQuote{of} as most other heads. Interestingly, these tokens also seem distinctive / descriptive for the given paragraph and the token \setQuote{email} which caused most perturbation in \cref{fig:prefix_perturb} is standing out.

\begin{figure*}[t]
 \centering
 \includegraphics[width=1.0\linewidth]{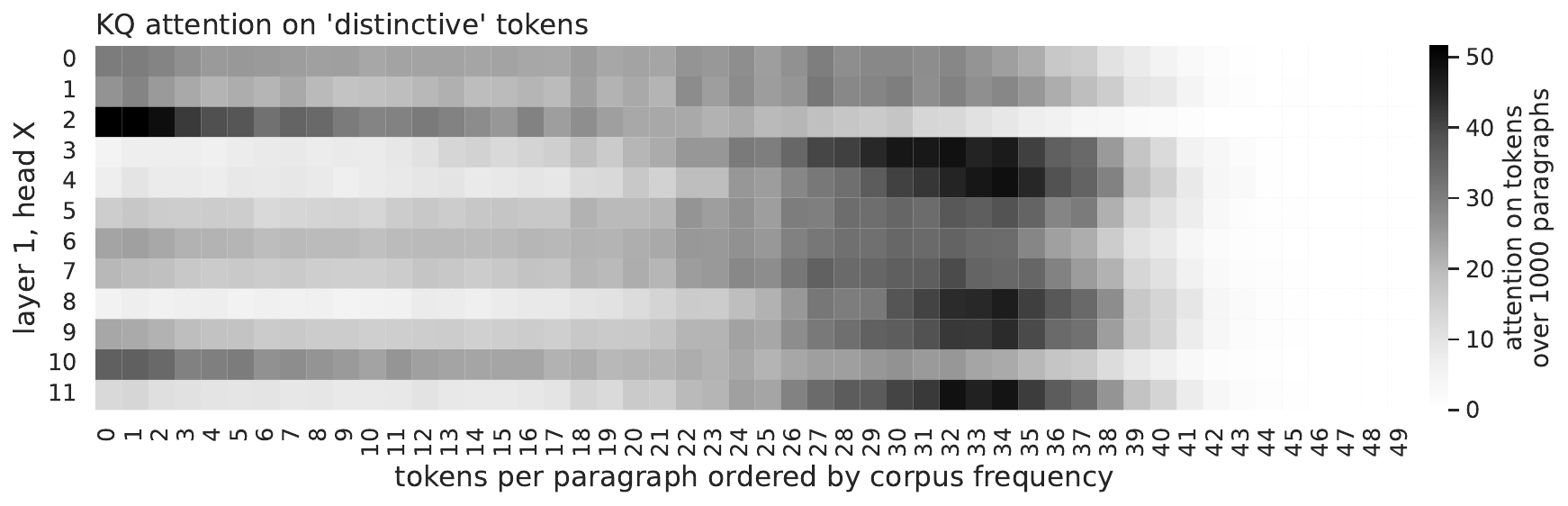} 
\caption{The \emph{memorization head} 2 in layer 1 is strongly negatively correlated (\num{-0.97}) with the corpus-level frequency of tokens. The plot shows the aggregated attention that each head assigns to tokens per paragraph ranked by corpus frequency. Note that, due to ties in token frequencies, often not all ranks up to rank \num{49} receive attention. 
}
\label{fig:rare_tokens}
\end{figure*}

\subsection{Activation Pattern Analysis}
\label{sec:activation_pattern}

We observe similar patterns when analyzing forward pass activations of \emph{key-query (KQ) attention patterns}. The normalized, inner product of \setQuote{keys} $\boldsymbol{k}$ and \setQuote{queries} $\boldsymbol{q}$ is given by $\softmax(\boldsymbol{k} \boldsymbol{q})$ and describes the amount of \setQuote{lookback} attention from the currently decoded token to all previous tokens. In our case, we choose to study the attention between the first decoded token onto the full \num{50}-token prefix as shown in \cref{fig:attention_pattern_l1} [bottom]. Similar to the activation gradients, head 2 attends to seemingly distinctive or rare tokens such as \setQuote{Subscribe}, \setQuote{Washington}, \setQuote{email} or \setQuote{offers} instead of more frequent tokens like punctuation marks and stop words as heads 3 to 11 do. Recent work \citep{tigges_linear_2023, sun_massive_2024} finds that punctuation marks often serve as \setQuote{aggregation points} for sentiment throughout a paragraph. It is important to note that these attention patterns per head look entirely different for any other layer, such as layer 2 visualized in \cref{fig:attention_pattern_l1_example2} in the appendix.

\subsection{Rare Token Correlation}
\label{sec:rare_token_correlation}

When perturbing tokens (\cref{sec:token_perturbation}), and analyzing activations (\cref{sec:activation_gradients}, \cref{sec:activation_pattern}), we find that \setQuote{rare} tokens play an important role for memorization, related to other previous findings on the relation between long tail distributions and memorization \citep{feldman_what_2020}. To test this \emph{rate token hypothesis}, we consider the unigram distribution of all tokens in our corpus which amounts to \num{34562} unique tokens. For every paragraph in our corpus, we rank the tokens by their corpus frequency from \num{0} (most rare) to \num{49} (most frequent) allowing ties. Then, we feed each paragraph to \gptNeoSmall, obtain the KQ attention of the first decoded token at onto every prefix token. We go through the paragraph's token frequency ranks and sum up the attention that each head assigns to the token of each rank. As shown in \cref{fig:rare_tokens}, we find that head number 2 in layer 1 is indeed the one most strongly correlated with rare tokens. As such, we have identified an important function of a model component that plays a vital role in memorizing paragraphs. One may hypothesize that the model computes a signature of each paragraph as a \setQuote{bag of its rare words}. It could then use this signature as a query to look up its \setQuote{memory of paragraphs} seen during training.

\section{Discussion}
\label{sec:discussion}

Our focus lies on identifying \setQuote{where} memorization-relevant model components may be localized, but our findings open up interesting follow-up questions on the \setQuote{why} and \setQuote{how}. In \cref{sec:editing}, we are unlearning and editing \MPs, but memorization may similarly lead to better performance or may be desired for certain types of paragraphs \citep{feldman_what_2020}. One could in fact take an opposite view and study how to make a model memorize an \NMP. Being able to identify differences in the model-internal processing of \MPs and \NMPs, future work could train a classifier on the activations or gradients \citep{pimentel_architectural_2022, li_inference-time_2023} to detect looming memorization at decoding time instead of considering logit distributions or post-hoc string matching \citep{shi_detecting_2023}. Similar to our token perturbations in \cref{sec:token_perturbation}, future work could attempt to divert memorized continuations through targeted interventions in the forward pass.

\section{Conclusion}
\label{sec:conlusion}

Gradients flow differently for memorized (more in lower layers) than for non-memorized paragraphs (more in higher layers). While many model components are involved, memorization is often localized to few, distinctive tokens in the prefix that are predominantly processed by the attention head 2 in layer 1 of \gptNeoSmall.

\section*{Acknowledgments}

We would like to thank the Google AI Developer Assistance team (AIDA) as well as Katherine Lee, Neel Nanda, Nicholas Carlini, Timo Denk, Richard Shin, Xiang Deng, Bin Ni, Alex Polozov, Luca Beurer-Kellner and Suchin Gururangan, and Mengzhou Xia.

\section*{Limitations}
\label{sec:limitations}

The purpose of this work is to study paragraph memorization of one model in detail. Our methodology is however not model-specific and can be applied to other models such as the Pythia family \citep{biderman_pythia_2023}. Another important direction is memorization in instruction- and RLHF-tuned models. Most prior work \citep{carlini_extracting_2020, carlini_quantifying_2022, mccoy_how_2023} and our paper identify memorization through prefix continuation, but instruction-tuned models may behave and memorize entirely differently. Importantly, there are ongoing discussions on the explanatory value of gradients \citep{sundararajan_axiomatic_2017, du_generalizing_2023} and activations \citep{farquhar_challenges_2023, stoehr_unsupervised_2024}. By combining different interpretability methods such as analyses of parameter gradients, activation gradients, token perturbation and patching, we make an effort to provide different perspectives and find that different methods point to similar model components and mechanisms. 

\section*{Impact Statement}

Language model memorization has important implications with respect to performance, copyright and privacy concerns. To limit risks, we specifically study a small, open-weight model \gptNeoSmall and a widely studied public training set. We hope that a better understanding of memorization can help improve model performance and promotes the open-sourcing of language models. Not wanting to publicly leak organization-internal data or risking copyright infringement is a primary blocker for open-source efforts.

\bibliography{references}

\begin{thebibliography}{38}
\expandafter\ifx\csname natexlab\endcsname\relax\def\natexlab#1{#1}\fi

\bibitem[{Biderman et~al.(2023)Biderman, Schoelkopf, Anthony, Bradley, O'Brien, Hallahan, Khan, Purohit, Prashanth, Raff, Skowron, Sutawika, and van~der Wal}]{biderman_pythia_2023}
Stella Biderman, Hailey Schoelkopf, Quentin Anthony, Herbie Bradley, Kyle O'Brien, Eric Hallahan, Mohammad~Aflah Khan, Shivanshu Purohit, USVSN~Sai Prashanth, Edward Raff, Aviya Skowron, Lintang Sutawika, and Oskar van~der Wal. 2023.
\newblock \href {https://doi.org/10.48550/ARXIV.2304.01373} {Pythia: {A} suite for analyzing large language models across training and scaling}.
\newblock \emph{arXiv}, 2304.01373.

\bibitem[{Brown et~al.(2020)Brown, Mann, Ryder, Subbiah, Kaplan, Dhariwal, Neelakantan, Shyam, Sastry, Askell, Agarwal, Herbert-Voss, Krueger, Henighan, Child, Ramesh, Ziegler, Wu, Winter, Hesse, Chen, Sigler, Litwin, Gray, Chess, Clark, Berner, McCandlish, Radford, Sutskever, and Amodei}]{brown_language_2020}
Tom Brown, Benjamin Mann, Nick Ryder, Melanie Subbiah, Jared~D Kaplan, Prafulla Dhariwal, Arvind Neelakantan, Pranav Shyam, Girish Sastry, Amanda Askell, Sandhini Agarwal, Ariel Herbert-Voss, Gretchen Krueger, Tom Henighan, Rewon Child, Aditya Ramesh, Daniel Ziegler, Jeffrey Wu, Clemens Winter, Chris Hesse, Mark Chen, Eric Sigler, Mateusz Litwin, Scott Gray, Benjamin Chess, Jack Clark, Christopher Berner, Sam McCandlish, Alec Radford, Ilya Sutskever, and Dario Amodei. 2020.
\newblock \href {https://proceedings.neurips.cc/paper/2020/file/1457c0d6bfcb4967418bfb8ac142f64a-Paper.pdf} {Language models are few-shot learners}.
\newblock In \emph{Advances in {Neural} {Information} {Processing} {Systems}}, volume~33, pages 1877--1901.

\bibitem[{Carlini et~al.(2022)Carlini, Ippolito, Jagielski, Lee, Tramer, and Zhang}]{carlini_quantifying_2022}
Nicholas Carlini, Daphne Ippolito, Matthew Jagielski, Katherine Lee, Florian Tramer, and Chiyuan Zhang. 2022.
\newblock \href {https://doi.org/10.48550/ARXIV.2202.07646} {Quantifying memorization across neural language models}.
\newblock \emph{ICLR}.

\bibitem[{Carlini et~al.(2020)Carlini, Tramer, Wallace, Jagielski, Herbert-Voss, Lee, Roberts, Brown, Song, Erlingsson, Oprea, and Raffel}]{carlini_extracting_2020}
Nicholas Carlini, Florian Tramer, Eric Wallace, Matthew Jagielski, Ariel Herbert-Voss, Katherine Lee, Adam Roberts, Tom Brown, Dawn Song, Ulfar Erlingsson, Alina Oprea, and Colin Raffel. 2020.
\newblock \href {https://doi.org/10.48550/ARXIV.2012.07805} {Extracting training data from large language models}.
\newblock \emph{arXiv}, 10.48550.

\bibitem[{Chang et~al.(2023)Chang, Thomason, and Jia}]{chang_localization_2023}
Ting-Yun Chang, Jesse Thomason, and Robin Jia. 2023.
\newblock \href {http://arxiv.org/abs/2311.09060} {Do localization methods actually localize memorized data in llms?}
\newblock ArXiv:2311.09060 [cs].

\bibitem[{Du et~al.(2023)Du, Torroba~Hennigen, Stoehr, Warstadt, and Cotterell}]{du_generalizing_2023}
Kevin Du, Lucas Torroba~Hennigen, Niklas Stoehr, Alex Warstadt, and Ryan Cotterell. 2023.
\newblock \href {https://doi.org/10.18653/v1/2023.acl-long.669} {Generalizing backpropagation for gradient-based interpretability}.
\newblock In \emph{{ACL}}, pages 11979--11995, Toronto, Canada.

\bibitem[{Eldan and Russinovich(2023)}]{eldan_whos_2023}
Ronen Eldan and Mark Russinovich. 2023.
\newblock \href {http://arxiv.org/abs/2310.02238} {Who's {Harry} {Potter}? {Approximate} unlearning in {LLMs}}.
\newblock ArXiv:2310.02238 [cs].

\bibitem[{Farquhar et~al.(2023)Farquhar, Varma, Kenton, Gasteiger, Mikulik, and Shah}]{farquhar_challenges_2023}
Sebastian Farquhar, Vikrant Varma, Zachary Kenton, Johannes Gasteiger, Vladimir Mikulik, and Rohin Shah. 2023.
\newblock \href {https://doi.org/10.48550/ARXIV.2312.10029} {Challenges with unsupervised {LLM} knowledge discovery}.
\newblock \emph{arXiv}, 2312.10029.

\bibitem[{Feldman and Zhang(2020)}]{feldman_what_2020}
Vitaly Feldman and Chiyuan Zhang. 2020.
\newblock \href {https://doi.org/10.48550/ARXIV.2008.03703} {What neural networks memorize and why: discovering the long tail via influence estimation}.
\newblock \emph{NeurIPS}.

\bibitem[{Gao et~al.(2021)Gao, Biderman, Black, Golding, Hoppe, Foster, Phang, He, Thite, Nabeshima, Presser, and Leahy}]{gao_pile_2021}
Leo Gao, Stella Biderman, Sid Black, Laurence Golding, Travis Hoppe, Charles Foster, Jason Phang, Horace He, Anish Thite, Noa Nabeshima, Shawn Presser, and Connor Leahy. 2021.
\newblock \href {https://doi.org/10.48550/ARXIV.2101.00027} {The {Pile}: {An} {800Gb} dataset of diverse text for language modeling}.
\newblock \emph{arXiv}, 2101.00027.

\bibitem[{Geva et~al.(2023)Geva, Bastings, Filippova, and Globerson}]{geva_dissecting_2023}
Mor Geva, Jasmijn Bastings, Katja Filippova, and Amir Globerson. 2023.
\newblock \href {https://doi.org/10.48550/ARXIV.2304.14767} {Dissecting recall of factual associations in auto-regressive language models}.
\newblock \emph{arXiv}, 2304.14767.

\bibitem[{Hartmann et~al.(2023)Hartmann, Suri, Bindschaedler, Evans, Tople, and West}]{hartmann_sok_2023}
Valentin Hartmann, Anshuman Suri, Vincent Bindschaedler, David Evans, Shruti Tople, and Robert West. 2023.
\newblock \href {http://arxiv.org/abs/2310.18362} {Sok: memorization in general-purpose large language models}.
\newblock ArXiv:2310.18362 [cs].

\bibitem[{Hase et~al.(2023)Hase, Bansal, Kim, and Ghandeharioun}]{hase_does_2023}
Peter Hase, Mohit Bansal, Been Kim, and Asma Ghandeharioun. 2023.
\newblock \href {https://doi.org/10.48550/ARXIV.2301.04213} {Does localization inform editing? {Surprising} differences in causality-based localization vs. knowledge editing in language models}.
\newblock \emph{NeurIPS}.

\bibitem[{Haviv et~al.(2023)Haviv, Cohen, Gidron, Schuster, Goldberg, and Geva}]{haviv_understanding_2023}
Adi Haviv, Ido Cohen, Jacob Gidron, Roei Schuster, Yoav Goldberg, and Mor Geva. 2023.
\newblock \href {https://doi.org/10.48550/ARXIV.2210.03588} {Understanding transformer memorization recall through idioms}.
\newblock \emph{EACL}.

\bibitem[{Hu et~al.(2021)Hu, Salcic, Sun, Dobbie, Yu, and Zhang}]{hu_membership_2021}
Hongsheng Hu, Zoran Salcic, Lichao Sun, Gillian Dobbie, Philip~S. Yu, and Xuyun Zhang. 2021.
\newblock \href {https://doi.org/10.48550/ARXIV.2103.07853} {Membership inference attacks on machine learning: {A} survey}.
\newblock \emph{ACM Computing Surveys}.

\bibitem[{Kingma and Ba(2015)}]{kingma_adam_2015}
Diederik Kingma and Jimmy Ba. 2015.
\newblock \href {https://arxiv.org/abs/1412.6980} {Adam: {A} method for stochastic optimization}.
\newblock In \emph{International {Conference} on {Learning} {Representations}}, page 337.

\bibitem[{Kirchenbauer et~al.(2023)Kirchenbauer, Geiping, Wen, Katz, Miers, and Goldstein}]{kirchenbauer_watermark_2023}
John Kirchenbauer, Jonas Geiping, Yuxin Wen, Jonathan Katz, Ian Miers, and Tom Goldstein. 2023.
\newblock \href {https://doi.org/10.48550/ARXIV.2301.10226} {A watermark for large language models}.
\newblock \emph{ICML}.

\bibitem[{Li et~al.(2023)Li, Patel, Viégas, Pfister, and Wattenberg}]{li_inference-time_2023}
Kenneth Li, Oam Patel, Fernanda Viégas, Hanspeter Pfister, and Martin Wattenberg. 2023.
\newblock \href {https://doi.org/10.48550/ARXIV.2306.03341} {Inference-time intervention: {Eliciting} truthful answers from a language model}.
\newblock \emph{NeurIPS}.

\bibitem[{Maini et~al.(2023)Maini, Mozer, Sedghi, Lipton, Kolter, and Zhang}]{maini_can_2023}
Pratyush Maini, Michael~C. Mozer, Hanie Sedghi, Zachary~C. Lipton, J.~Zico Kolter, and Chiyuan Zhang. 2023.
\newblock \href {https://doi.org/10.48550/ARXIV.2307.09542} {Can neural network memorization be localized?}
\newblock \emph{ICML}.

\bibitem[{Mattern et~al.(2023)Mattern, Mireshghallah, Jin, Schoelkopf, Sachan, and Berg-Kirkpatrick}]{mattern_membership_2023}
Justus Mattern, Fatemehsadat Mireshghallah, Zhijing Jin, Bernhard Schoelkopf, Mrinmaya Sachan, and Taylor Berg-Kirkpatrick. 2023.
\newblock \href {https://doi.org/10.18653/v1/2023.findings-acl.719} {Membership inference attacks against language models via neighbourhood comparison}.
\newblock In \emph{Findings of {ACL}}, pages 11330--11343.

\bibitem[{McCoy et~al.(2023)McCoy, Smolensky, Linzen, Gao, and Celikyilmaz}]{mccoy_how_2023}
Thomas McCoy, Paul Smolensky, Tal Linzen, Jianfeng Gao, and Asli Celikyilmaz. 2023.
\newblock \href {https://doi.org/10.1162/tacl_a_00567} {How much do language models copy from their training data? {Evaluating} linguistic novelty in text generation using raven}.
\newblock \emph{Transactions of the Association for Computational Linguistics}, 11:652--670.

\bibitem[{Meng et~al.(2022)Meng, Bau, Andonian, and Belinkov}]{meng_locating_2022}
Kevin Meng, David Bau, Alex Andonian, and Yonatan Belinkov. 2022.
\newblock \href {https://doi.org/10.48550/ARXIV.2202.05262} {Locating and editing factual associations in {GPT}}.
\newblock \emph{NeurIPS}.

\bibitem[{Nanda(2023)}]{nanda_transformerlenslibrary_2023}
Neel Nanda. 2023.
\newblock \href {https://github. com/neelnanda-io/TransformerLens} {{TransformerLens}—{A} library for mechanistic interpretability of generative language models}.

\bibitem[{Nasr et~al.(2023)Nasr, Carlini, Hayase, Jagielski, Cooper, Ippolito, Choquette-Choo, Wallace, Tramèr, and Lee}]{nasr_scalable_2023}
Milad Nasr, Nicholas Carlini, Jonathan Hayase, Matthew Jagielski, A.~Feder Cooper, Daphne Ippolito, Christopher~A. Choquette-Choo, Eric Wallace, Florian Tramèr, and Katherine Lee. 2023.
\newblock \href {http://arxiv.org/abs/2311.17035} {Scalable extraction of training data from (production) language models}.

\bibitem[{{New York Times}(2023)}]{new_york_times_one_2023}
{New York Times}. 2023.
\newblock \href {https://nytco-assets.nytimes.com/2023/12/Lawsuit-Document-dkt-1-68-Ex-J.pdf} {One hundred examples of {GPT}-4 memorizing content from the {New} {York} {Times}, {Document} 1-68, {Exhibit} {J}}.

\bibitem[{Pal et~al.(2023)Pal, Sun, Yuan, Wallace, and Bau}]{pal_future_2023}
Koyena Pal, Jiuding Sun, Andrew Yuan, Byron~C. Wallace, and David Bau. 2023.
\newblock \href {https://doi.org/10.48550/ARXIV.2311.04897} {Future {Lens}: {Anticipating} subsequent tokens from a single hidden state}.
\newblock \emph{CoNLL}.

\bibitem[{Pimentel et~al.(2022)Pimentel, Valvoda, Stoehr, and Cotterell}]{pimentel_architectural_2022}
Tiago Pimentel, Josef Valvoda, Niklas Stoehr, and Ryan Cotterell. 2022.
\newblock \href {https://arxiv.org/pdf/2211.06420.pdf} {The architectural bottleneck principle}.
\newblock In \emph{Conference on {Empirical} {Methods} in {Natural} {Language} {Processing} ({EMNLP})}.

\bibitem[{Power et~al.(2022)Power, Burda, Edwards, Babuschkin, and Misra}]{power_grokking_2022}
Alethea Power, Yuri Burda, Harri Edwards, Igor Babuschkin, and Vedant Misra. 2022.
\newblock \href {https://doi.org/10.48550/ARXIV.2201.02177} {Grokking: {Generalization} beyond overfitting on small algorithmic datasets}.
\newblock \emph{arXiv}, 2201.02177.

\bibitem[{Shi et~al.(2023)Shi, Ajith, Xia, Huang, Liu, Blevins, Chen, and Zettlemoyer}]{shi_detecting_2023}
Weijia Shi, Anirudh Ajith, Mengzhou Xia, Yangsibo Huang, Daogao Liu, Terra Blevins, Danqi Chen, and Luke Zettlemoyer. 2023.
\newblock \href {https://doi.org/10.48550/ARXIV.2310.16789} {Detecting pretraining data from large language models}.
\newblock \emph{arXiv}, 2310.16789.

\bibitem[{Stoehr et~al.(2024)Stoehr, Cheng, Wang, Preotiuc-Pietro, and Bhowmik}]{stoehr_unsupervised_2024}
Niklas Stoehr, Pengxiang Cheng, Jing Wang, Daniel Preotiuc-Pietro, and Rajarshi Bhowmik. 2024.
\newblock \href {https://arxiv.org/abs/2309.06991} {Unsupervised contrast-consistent ranking with language models}.
\newblock \emph{Proceedings of the 18th Conference of the European Chapter of the Association for Computational Linguistics}.

\bibitem[{Sun et~al.(2024)Sun, Chen, Kolter, and Liu}]{sun_massive_2024}
Mingjie Sun, Xinlei Chen, J.~Zico Kolter, and Zhuang Liu. 2024.
\newblock \href {https://doi.org/10.48550/ARXIV.2402.17762} {Massive activations in large language models}.
\newblock \emph{arXiv}, 2402.17762.

\bibitem[{Sundararajan et~al.(2017)Sundararajan, Taly, and Yan}]{sundararajan_axiomatic_2017}
Mukund Sundararajan, Ankur Taly, and Qiqi Yan. 2017.
\newblock \href {https://arxiv.org/pdf/1703.01365.pdf} {Axiomatic attribution for deep networks}.
\newblock In \emph{{ICML}}, pages 3319--3328.
\newblock Event-place: Sydney, NSW, Australia.

\bibitem[{Tigges et~al.(2023)Tigges, Hollinsworth, Geiger, and Nanda}]{tigges_linear_2023}
Curt Tigges, Oskar~John Hollinsworth, Atticus Geiger, and Neel Nanda. 2023.
\newblock \href {http://arxiv.org/abs/2310.15154} {Linear representations of sentiment in large language models}.
\newblock \emph{arXiv}, 2310.15154.

\bibitem[{Vaswani et~al.(2017)Vaswani, Shazeer, Parmar, Uszkoreit, Jones, Gomez, Kaiser, and Polosukhin}]{vaswani_attention_2017}
Ashish Vaswani, Noam Shazeer, Niki Parmar, Jakob Uszkoreit, Llion Jones, Aidan~N. Gomez, Lukasz Kaiser, and Illia Polosukhin. 2017.
\newblock \href {https://papers.nips.cc/paper/2017/hash/3f5ee243547dee91fbd053c1c4a845aa-Abstract.html} {Attention is all you need}.
\newblock In \emph{{NeurIPS}}.

\bibitem[{Yu et~al.(2023)Yu, Merullo, and Pavlick}]{yu_characterizing_2023}
Qinan Yu, Jack Merullo, and Ellie Pavlick. 2023.
\newblock \href {https://doi.org/10.48550/ARXIV.2310.15910} {Characterizing mechanisms for factual recall in language models}.
\newblock \emph{arXiv}, 2310.15910.

\bibitem[{Zhang et~al.(2021)Zhang, Ippolito, Lee, Jagielski, Tramèr, and Carlini}]{zhang_counterfactual_2021}
Chiyuan Zhang, Daphne Ippolito, Katherine Lee, Matthew Jagielski, Florian Tramèr, and Nicholas Carlini. 2021.
\newblock \href {https://doi.org/10.48550/ARXIV.2112.12938} {Counterfactual memorization in neural language models}.
\newblock \emph{NeurIPS}.

\bibitem[{Zhang and Nanda(2024)}]{zhang_towards_2024}
Fred Zhang and Neel Nanda. 2024.
\newblock \href {https://paperswithcode.com/paper/towards-best-practices-of-activation-patching} {Towards best practices of activation patching in language models: {Metrics} and methods}.
\newblock In \emph{{ICLR}}.

\bibitem[{Zheng and Jiang(2022)}]{zheng_empirical_2022}
Xiaosen Zheng and Jing Jiang. 2022.
\newblock \href {https://doi.org/10.18653/v1/2022.acl-long.434} {An empirical study of memorization in {NLP}}.
\newblock In \emph{{ACL}}, pages 6265--6278, Dublin, Ireland.

\end{thebibliography}

\appendix

\newpage
\section{Appendix}
\label{sec:appendix}

\subsection{Paragraph Pre-Processing}
\label{sec:preprocessing}

We filter out paragraphs containing any variation of keywords that are disproportionally frequent in \citet{carlini_quantifying_2022}'s \href{https://github.com/ethz-spylab/lm_memorization_data/tree/main/data}{\pile subset}: \setQuote{TripAdvisor, href, license, copyright, software, manuscript, submission, distribution, disclaimed, limited}. As a second preprocessing step, we filter out all paragraphs that contain less than \num{50}\% of unique tokens to remove paragraphs containing mostly white spaces. 

\subsection{Activation Analysis at Selected Tokens}
\label{sec:perturbed_activations}

\begin{figure*}[t]
 \centering
 \includegraphics[width=1.0\linewidth]{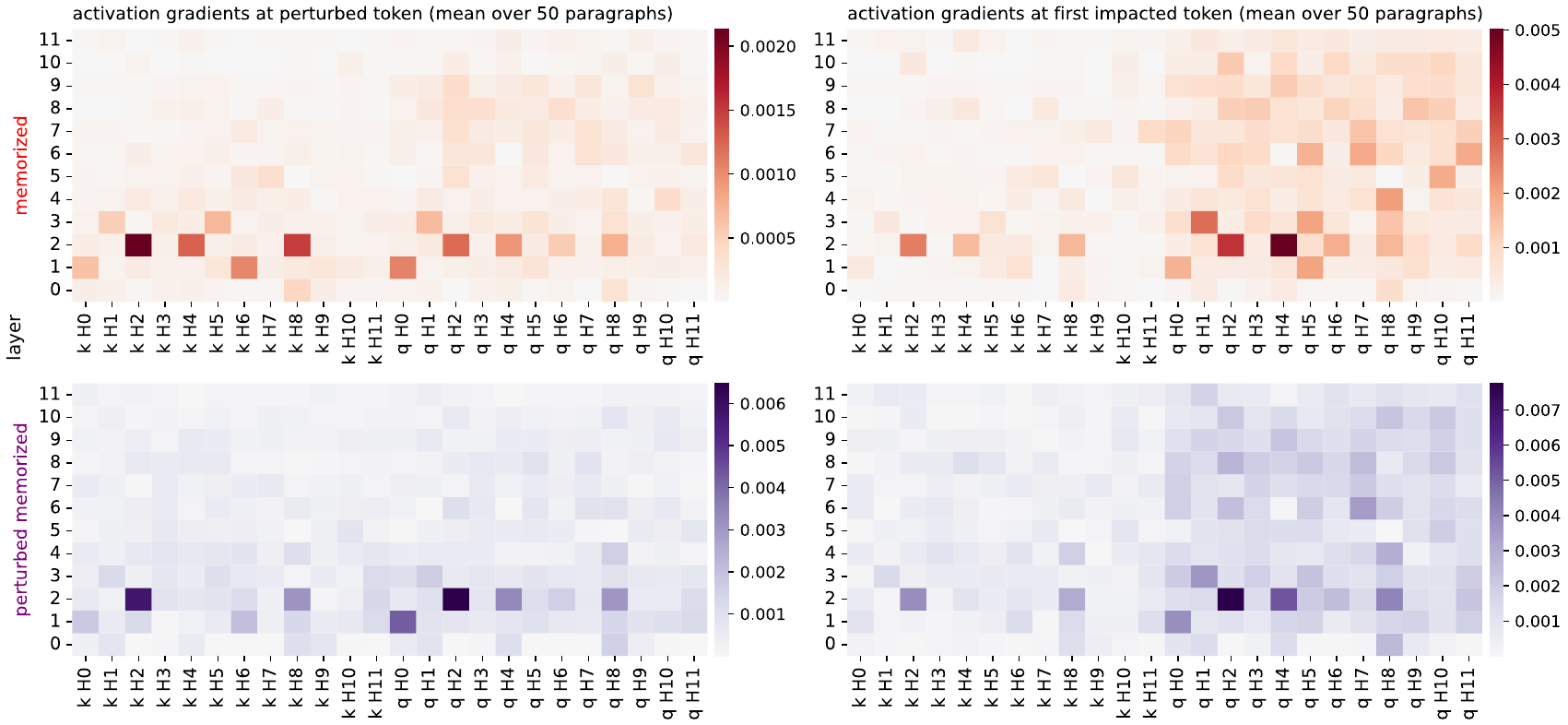} 
\caption{Comparing activation gradients of \num{50} \memParas and their \perturbmemColor{perturbed memorized} counterparts at the perturbed token and the first impacted token.}
\label{fig:activ_grads_perturbed}
\end{figure*}

In \cref{sec:token_perturbation}, we perturb single tokens in the prefix and measure the incurred change in the model's continuation with respect to the originally, memorized continuation. We then pick the token position that causes the maximum change and term it the \emph{perturbed token}. In the model's generation, we pick the first token that is changed with respect to the unperturbed continuation and call it the \emph{impact token}. Next, we pass both paragraphs, the \memPara and the \perturbmemPara, to the model and compute the activation gradients at the perturbed and the impact token following \cref{sec:activation_gradients}. The result in \cref{fig:activ_grads_perturbed} shows large gradients for key and query activations at layer 2. At the impacted token, query activation gradients are generally more active.

\begin{table*}[t]
\fontsize{10}{10}\selectfont
\centering
\renewcommand{\arraystretch}{1.4} 
\setlength{\tabcolsep}{0.35em} 
\begin{tabular}{c|c|l}
NLL & EM  & paragraph                                                                                       \\ \hline
0.426 & 50 & \scriptsize ;Classes</a></li>
</ul>
<div>
<script type="text/javascript"><!--
  allClassesLink = document.getElementById("allclasses\_navbar\_top");
  if\ldots
\\
1.074 & 50 & \scriptsize  headlines out of Washington never seem to slow. Subscribe to The D.C. Brief to make sense of what matters most. Please enter a valid\ldots \\
0.387 & 50 & \scriptsize  Sign up for Take Action Now and get three actions in your inbox every week. You will receive occasional promotional offers for \ldots\\
0.259 & 50 & \scriptsize The following are trademarks or service marks of Major League Baseball entities and may be used only with permission of Major League\ldots\\
0.276 & 50 & \scriptsize 0" SUMMARY="">\escape{n}<TR>\escape{n}<TD COLSPAN=2 BGCOLOR="\#EEEEFF" CLASS="NavBarCell1">\escape{n}<A NAME="navbar\_top\_firstrow"\dots
\end{tabular}
\caption{Representative paragraphs that are memorized by \gptNeoSmall based on the exact match (\EM) of \num{50} tokens between the model's generation and the ground truth training set paragraph.}
\label{tab:memorized_paragraphs}
\end{table*}

\begin{table*}[t]
\fontsize{8}{8}\selectfont
\centering
\renewcommand{\arraystretch}{1.4} 
\setlength{\tabcolsep}{0.35em} 
\begin{tabular}{p{5cm}|p{5cm}|p{5cm}}
\textbf{prefix}                                                                                                                                                                                                                                                                                                                                       & \textbf{memorized continuation}                                                                                                                                                                                                                                                                                                      & \textbf{perturbed continuation}                                                                                                                                                                                                                                                                                           \\ \hline
\textgreater{}\textless{}/li\textgreater{}\textbackslash{}n\textless{}/ul\textgreater{}\textbackslash{}n\textless{}div\textgreater{}\textbackslash{}n\textless{}script type="text/javascript"\textgreater{}\textless{}!--\textbackslash{}n  allClassesLink = document.getElementById("allclasses\_navbar \_bottom");\textbackslash{}n  if(window==top & ) \{\textbackslash{}n    allClassesLink.style.display = "block";\textbackslash{}n  \}\textbackslash{}n  else \{\textbackslash{}n    allClassesLink.style.display = "none";\textbackslash{}n  \}\textbackslash{}n  //--\textgreater{}\textbackslash{}n\textless{}/script                                                              & = "\&\#x2F; top";\textbackslash{}n  visibility = \{ Tosink: "visibility" \};\textbackslash{}n adaption = document.getElementById("adaption");\textbackslash{}n  if(idaption)\textbackslash{}n    aada                                                                                                                     \\ \hline
headlines out of Washington never seem to slow. Subscribe to The D.C. Brief to make sense of what matters most. Please enter a valid email address. Sign Up Now Check the box if you do not wish to receive promotional offers via email from                                                                                                         & TIME. You can unsubscribe at any time. By signing up you are agreeing to our Terms of Use and Privacy Policy . This site is protected by reCAPTCHA and the Google Privacy Policy and Terms of Service apply. Thank you! For your                                                                                                     & icon.com/dccbrief./\textbackslash{}n/\textbackslash{}nThe D.C. Brief is a weekly newsletter focused on the latest news and information from the Washington, D.C. area. Sign up here./\textbackslash{}n/\textbackslash{}nThe D.C. Brief                                                                                    \\ \hline
Sign up for Take Action Now and get three actions in your inbox every week. You will receive occasional promotional offers for programs that support The Nation’s journalism. You can read our Privacy Policy here. Sign up for Take Action Now and get                                                                                               & three actions in your inbox every week.\textbackslash{}n\textbackslash{}nThank you for signing up. For more from The Nation, check out our latest issue\textbackslash{}n\textbackslash{}nSubscribe now for as little as \$2 a month!\textbackslash{}n\textbackslash{}nSupport Progressive Journalism The Nation is reader supported: & 03 in your inbox every week.\textbackslash{}n\textbackslash{}nThank you for signing up. For more from The Nation, check out our latest issue\textbackslash{}n\textbackslash{}nSubscribe now for as little as \$2 a month!\textbackslash{}n\textbackslash{}nSupport Progressive Journalism The Nation is reader supported: \\ \hline
The following are trademarks or service marks of Major League Baseball entities and may be used only with permission of Major League Baseball Properties, Inc. or the relevant Major League Baseball entity: Major League, Major League Baseball, MLB, the silhouetted batter logo                                                                    & , World Series, National League, American League, Division Series, League Championship Series, All-Star Game, and the names, nicknames, logos, uniform designs, color combinations, and slogans designating the Major League Baseball clubs and entities, and                                                                        & and/or othershaped breakdown may loom, featuring provisions of the Koehlers and/or its logo, and may include design events or other references to the Koehlers and/or their logo.\textbackslash{}n\textbackslash{}nYamaha,                                                                                                \\ \hline
0" SUMMARY=""\textgreater{}\textbackslash{}n\textless{}TR\textgreater{}\textbackslash{}n\textless{}TD COLSPAN=2 BGCOLOR="\#EEEEFF" CLASS="NavBarCell1"\textgreater{}\textbackslash{}n\textless{}A NAME="navbar/\_top/\_firstrow"\textgreater{}\textless{}!-- --                                                                                       & \textgreater{}\textless{}/A\textgreater{}\textbackslash{}n\textless{}TABLE BORDER="0" CELLPADDING="0" CELLSPACING="3" SUMMARY=""\textgreater{}\textbackslash{}n  \textless{}TR ALIGN="center" VALIGN="top"\textgreater{}\textbackslash{}n  \textless{}TD                                                                             & TD\textgreater{}\textbackslash{}n\textless{}TD VALIGN=3\textgreater{}\textless{}FONT SIZE="-2"\textgreater{}\textbackslash{}n\&nbsp;PREV\&nbsp;NEXT\textless{}/FONT \textgreater  \textless{}/TD\textgreater{}\textbackslash{}n\textless{}TD VALIGN=3\textgreater{}\textless{}FONT SIZE="                                 
\end{tabular}
\caption{In \cref{sec:token_perturbation}, we perturb tokens in the prefix \textbf{[left]} and check how the model's perturbed continuations \textbf{[right]} changes with respect to the original, memorized continuation \textbf{[center]}. We find that perturbed continuations are still largely syntactically and semantically valid.}
\label{tab:perturbed_continuations}
\end{table*}

\begin{figure*}[h!]
 \centering
 \includegraphics[width=0.49\linewidth]{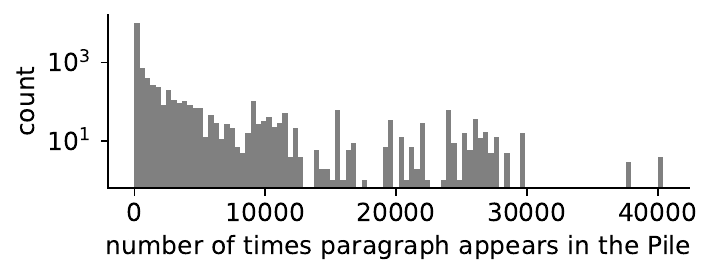} 
 \hfill
 \includegraphics[width=0.49\linewidth]{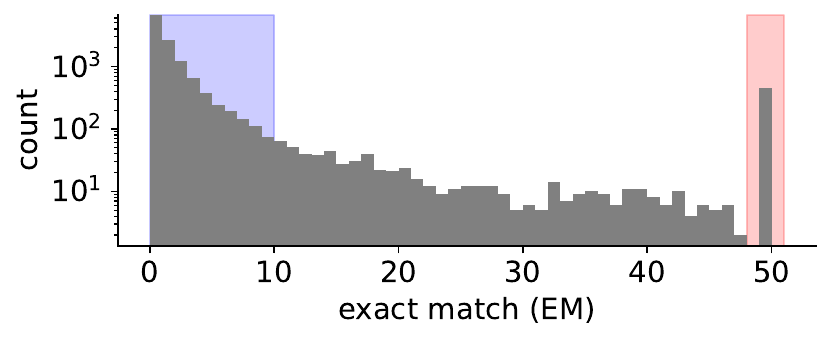} 
\caption{\textbf{[left]} Frequency count of each paragraph in our \pile subset borrowed from \citet{carlini_quantifying_2022}. \textbf{[right]} Exact match (\EM) distribution for all paragraphs in our dataset. We consider paragraphs with $\EM=50$ as \memColor{memorized} and $0 \leq \EM \leq 50$ as \nonmemColor{non-memorized}.}
\label{fig:data_stats}
\end{figure*}

\begin{figure*}[t]
 \centering
 \includegraphics[width=1.0\linewidth]{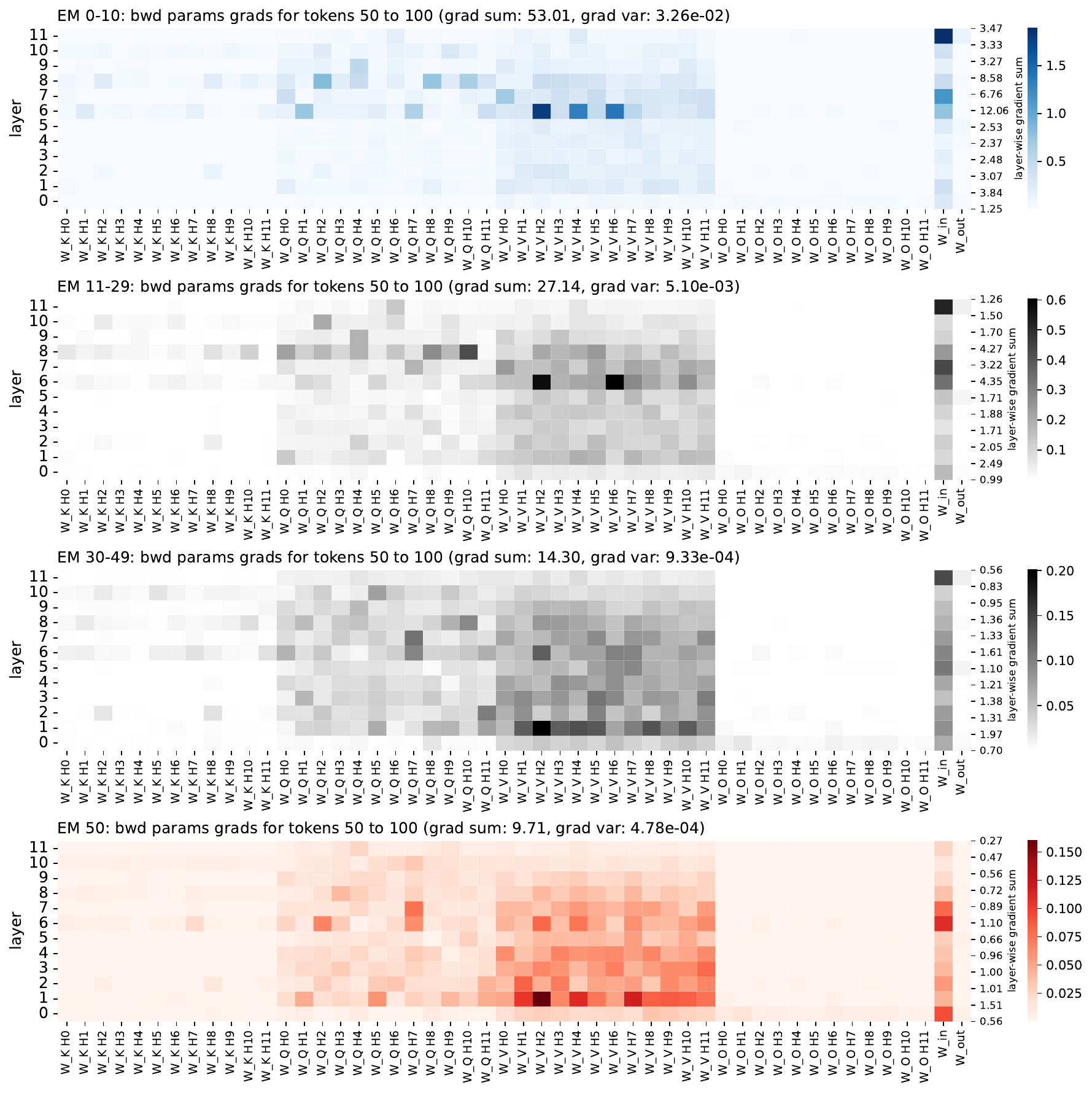} 
 \includegraphics[width=1.0\linewidth]{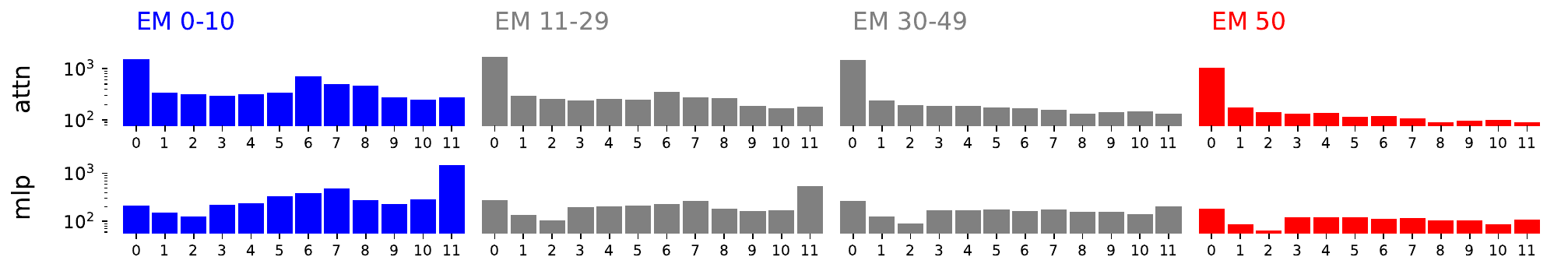} 
\caption{Supplementary plot to \cref{fig:params_gradients} showing the parameter gradients for paragraphs with \nonmemColor{$0 \leq \EM \leq 10$}, $11 \leq \EM \leq 29$, $30 \leq \EM \leq 49$ and \memColor{$\EM = 50$}. The bar plots visualize the absolute gradient sums for all attention (\textrm{attn}) and multi-layer perceptron (\textrm{mlp}) blocks. For \memParas, we observe that overall gradient flower is less and lower layers tend to have higher gradients. This change in gradient patterns from \nonmemParas to \memParas appears to be smooth.}
\label{fig:params_gradient_4}
\end{figure*}

\begin{figure*}[t]
 \centering
 \includegraphics[width=0.49\linewidth]{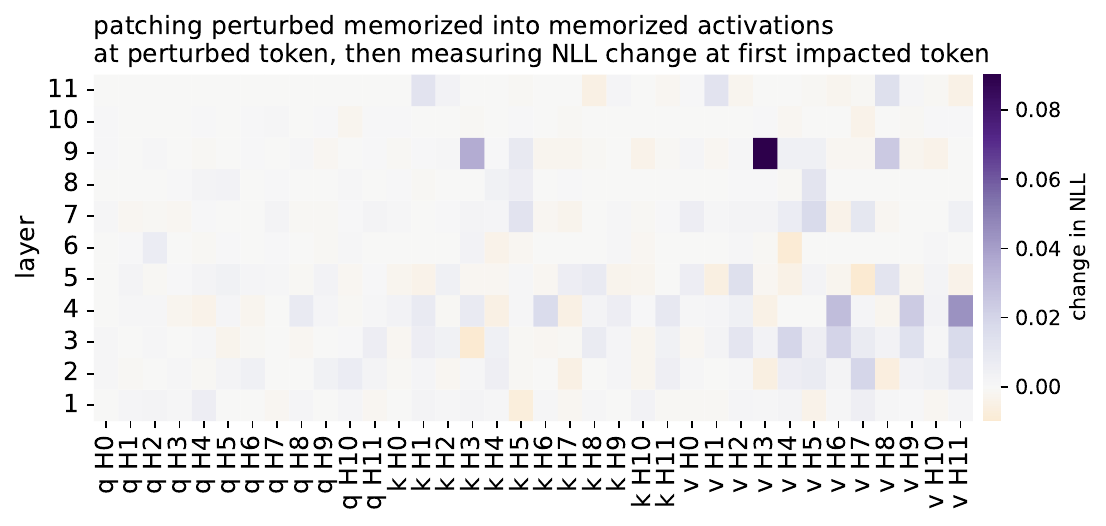} 
 \hfill
  \includegraphics[width=0.49\linewidth]{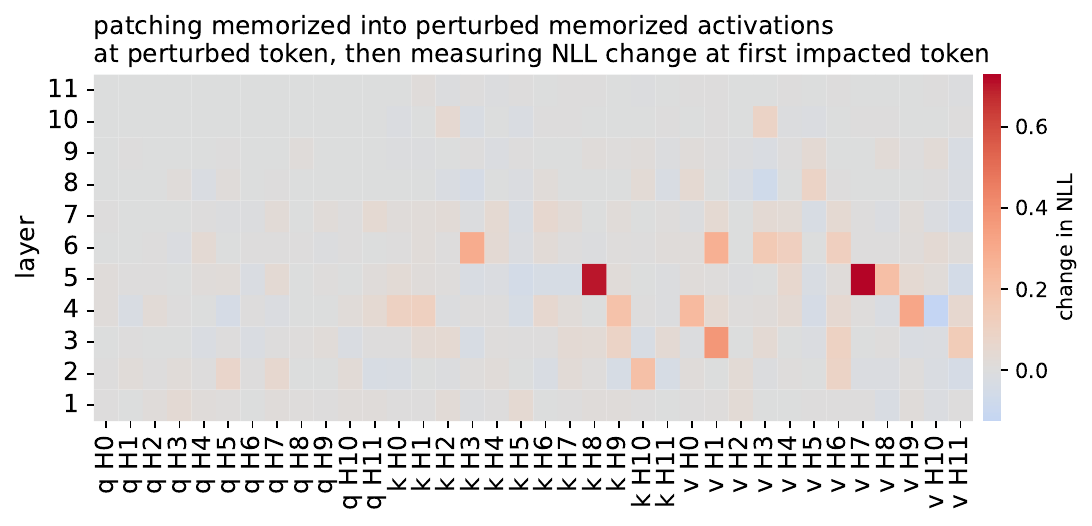} 
\caption{\setQuote{Two-way activation patching} at the perturbed token to identify the change in the impacted token.}
\label{fig:activ_patching}
\end{figure*}

\subsection{Patching-based Attribution}
\label{sec:patching}

In addition to studying activation gradients at the perturbed and impacted token, we experiment with activation patching \citep{meng_locating_2022, pal_future_2023, zhang_towards_2024}. We either consider \perturbmemParas as the \emph{clean run} and patch in activations at the perturbed token position from the \memParas or vice versa. As a patching metric, we measure the change in \NLL at the first impacted token. The results for \num{50} different paragraphs are presented in \cref{fig:activ_patching}.

\begin{figure*}[t]
 \centering
 \includegraphics[width=1.0\linewidth]{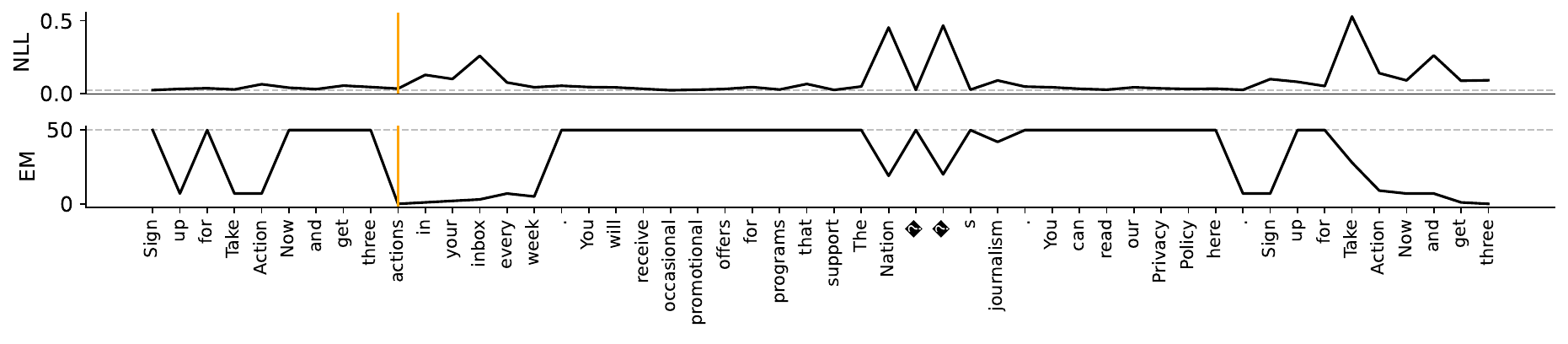} 
 \includegraphics[width=1.0\linewidth]{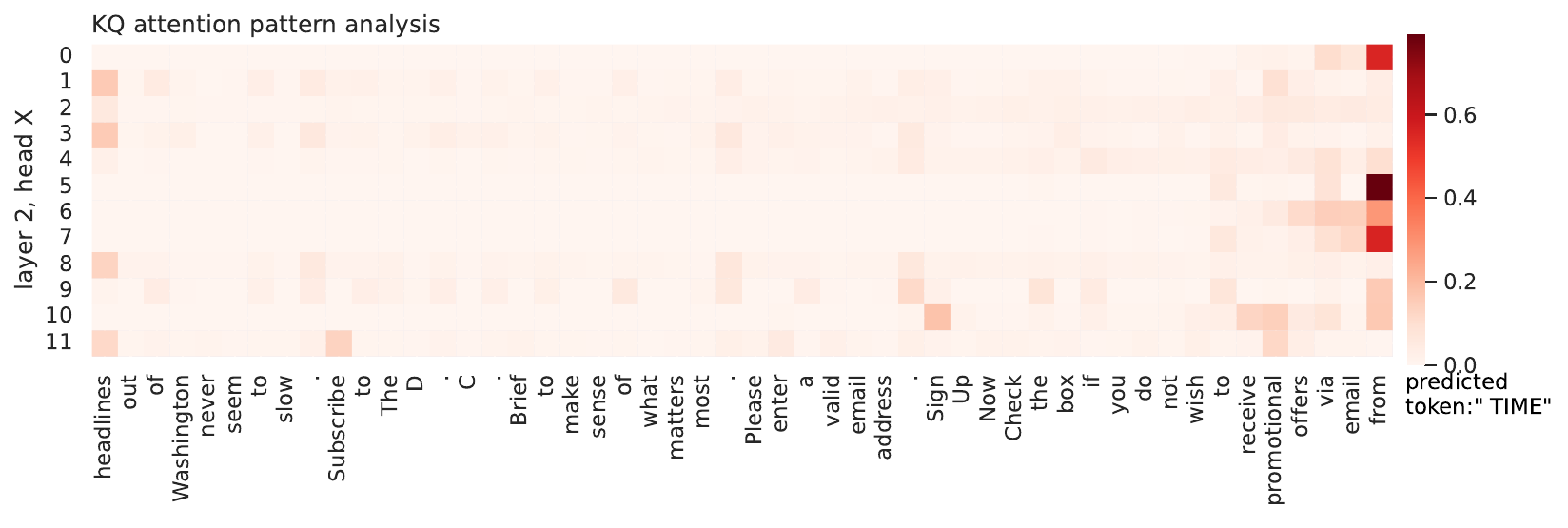} 
\caption{\textbf{[top]} Another example of perturbing the prefix tokens of a \memPara as presented in \cref{fig:prefix_perturb}. \textbf{[bottom]} Analysis if KQ attention patterns on layer \num{2} to compare against patterns in layer 1 presented in \cref{fig:attention_pattern_l1}.}
\label{fig:attention_pattern_l1_example2}
\end{figure*}

\end{document}